\documentclass[11pt]{article}

\usepackage[final]{acl}

\usepackage{times}
\usepackage{latexsym}

\usepackage[T1]{fontenc}

\usepackage[utf8]{inputenc}

\usepackage{microtype}

\usepackage{inconsolata}

\usepackage{graphicx}

\usepackage{microtype}
\usepackage{booktabs,arydshln}
\usepackage{array}
\usepackage{ragged2e}
\usepackage{amsmath}
\usepackage{threeparttable}
\usepackage{multicol}
\usepackage{multirow}
\usepackage{ccicons}
\usepackage{chngcntr}

\usepackage{afterpage}

\usepackage[capitalize]{cleveref}
\crefname{figure}{Fig.}{Figs.}
\crefname{table}{Tab.}{Tabs.}
\crefname{appendix}{App.}{Apps.}

\usepackage{booktabs}   
\usepackage{colortbl}   
\usepackage[table]{xcolor} 
\usepackage[table,HTML]{xcolor}

\usepackage{todonotes}
\usepackage{soul}

\definecolor{TodoColor}{rgb}{1,0.7,0.6}

\usepackage{enumitem}
\usepackage{placeins}
\newcommand{\huggingfacesmall}{\includegraphics[width=9px]{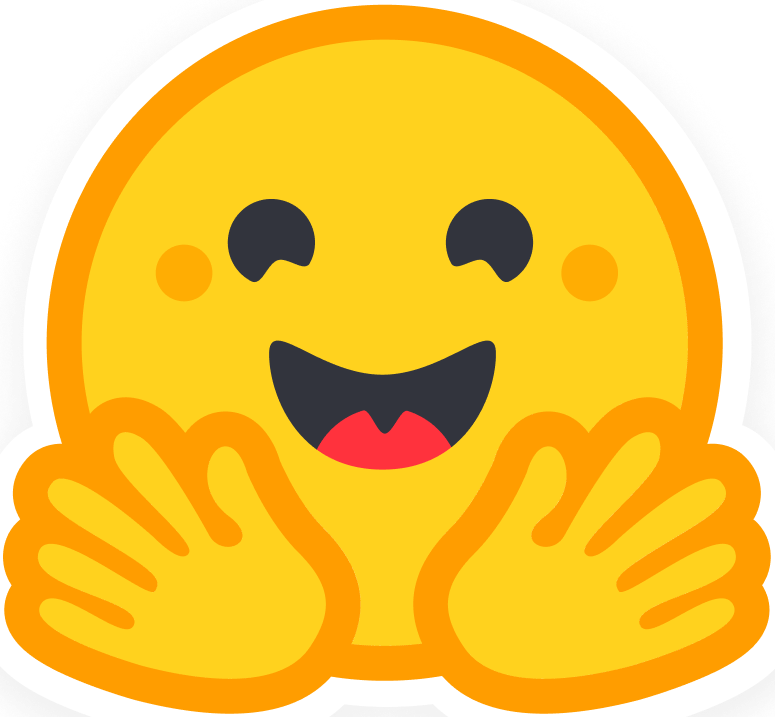}}

\usepackage{booktabs}
\usepackage{array}
\usepackage{arydshln}

\makeatletter
\def\adl@drawiv#1#2#3{%
        \hskip.5\tabcolsep
        \xleaders#3{#2.5\@tempdimb #1{1}#2.5\@tempdimb}%
                #2\z@ plus1fil minus1fil\relax
        \hskip.5\tabcolsep}
\newcommand{\cdashlinelr}[1]{%
  \noalign{\vskip\aboverulesep
           \global\let\@dashdrawstore\adl@draw
           \global\let\adl@draw\adl@drawiv}
  \cdashline{#1}
  \noalign{\global\let\adl@draw\@dashdrawstore
           \vskip\belowrulesep}}
\makeatother

\usepackage{amssymb}
\usepackage{pifont}

\usepackage{CJKutf8}
\usepackage{array}
\usepackage{multirow}
\newcommand{\zh}[1]{\begin{CJK}{UTF8}{gbsn}#1\end{CJK}}
\usepackage{threeparttable}
\title{Multilingual Long-Form Speech Instruction Following:\\
KIT's Submission to IWSLT 2026}

\author{ Enes Yavuz Ugan$^{1}$, Maike Züfle$^{1}$, Yuka Ko$^{1}$, Supriti Sinhamahapatra$^{1}$, \\
\bf{Fabian Retkowski$^{1}$, Seymanur Akti$^{1}$, Jan Niehues$^{1}$, Alexander Waibel$^{1,2}$} \\
$^{1}$Karlsruhe Institute of Technology \\
$^{2}$Carnegie Mellon University \\
firstname.lastname@kit.edu}

\begin{document}
\maketitle
\begin{abstract}
With the advent of Large Language Models, single-task and token-based multi-task models have evolved into instruction-based systems that infer task and target language implicitly from natural language prompts. This trend is reflected in IWSLT's Instruction Following Track, which this year introduced new tasks including an unknown surprise task, posing a genuine challenge against overfitting to known tasks. We present KIT's submission to the Long and Short Instruction Following tracks in the unconstrained setting. Our approach combines a general data augmentation pipeline that converts short-form corpora into long-form training data through segment concatenation, LLM-based label generation, and cross-lingual translation, yielding over 1M instances across six tasks and four languages. We further show that likelihood-based re-ranking, while highly effective for ASR, systematically degrades semantic tasks by spuriously selecting candidates generated from segmented audio processing rather than holistic long-form inference, a failure mode resolved by combining likelihood with Minimum Bayes Risk decoding.

\end{abstract}
\section{Introduction}
Recent work has focused on integrating speech into LLMs to create Speech Language Models, typically by incorporating pre-trained audio encoders \cite{tangsalmonn, koneru2025kit, retkowski-etal-2025-summarizing, zufle-niehues-2025-contrastive} into the LLM architecture. Alternative approaches train audio and vision encoders jointly with an LLM backbone, developing true multimodal Foundation Models \cite{Qwen2.5-Omni}, with recent work also demonstrating the effectiveness of combining speech and vision modalities \cite{sinhamahapatra2025slides,koneru2026boom}. Despite these advances, a significant gap remains: the effective processing of long-form audio \cite{papi2026mcif}. Most models rely on the Whisper encoder \cite{radford2022robustspeechrecognitionlargescale}, which natively supports only 30 seconds per inference pass. While newer models such as Phi-4 \cite{abdin2024phi} and Qwen2.5-Omni \cite{xu2025qwen25omnitechnicalreport} remove these architectural constraints, they lack exposure to long-form audio during training, a gap that manifests as significant performance degradation even on basic ASR \cite{papi2026mcif}.
This paper presents KIT's submission to the \href{https://iwslt.org/2026/instruction-following}{Unconstrained Long \& Short Instruction Following tracks of IWSLT 2026}. We participate in all four target languages: German, English, Italian, and Chinese. Our approach combines data augmentation to extend short-form datasets to long-form settings, temperature-scaled interleaving to balance task representation, and re-ranking to improve generation quality, exploring both end-to-end and cascaded architectures.
The main contributions of this work are:
\begin{itemize}
\item A data augmentation framework that includes conversion of short-form speech datasets into long-form instruction-following training data through segment concatenation, LLM-based label generation, and cross-lingual reference translation, yielding a publicly released dataset of over 1M instances across six tasks and four languages.\footnote{\huggingfacesmall{}\href{https://huggingface.co/datasets/YapayNet/iwslt2026-if-augmented}{YapayNet/iwslt2026-if-augmented}}
\item An empirical comparison of fixed-probability and temperature-scaled data interleaving strategies, identifying $T=2$ as a strong choice for multimodal speech instruction following.
\item A negative result on Chain-of-Thought task-token
conditioning, showing that prefix-based task routing
fails under task imbalance and task similarity,
leading to a collapse in task discrimination.
\item A systematic comparison of six re-ranking strategies under the realistic constraint of no task identity at inference time, revealing a previously uncharacterized failure mode where likelihood-based re-ranking spuriously selects segmentation-based candidates for semantic tasks.
\item A combined Likelihood+MBR re-ranking strategy that resolves the ASR-vs-semantics tradeoff, achieving strong ASR improvement while limiting degradation on QA and summarization.
\end{itemize}
\section{Data}
\label{sec:data}
This year's IWSLT instruction-following task covered six tasks: Automatic Speech Recognition (ASR), Speech Translation (ST), Spoken Question Answering (SQA), Speech Summarization (SSUM), Audio Chaptering (ACHAP), and a surprise task. All tasks except ASR support the language pairs en–en, de, it, zh; however, not all task–language combinations have in-domain data readily available. A further challenge is data format: most existing datasets consist of short utterances under 30 seconds, while the long-form track requires audio up to 15 minutes. We address both gaps through a three-stage augmentation framework: (1) segment concatenation with speaker-aware grouping to construct long-form audio, (2) LLM-based label generation to create task annotations for unlabeled or partially annotated data, and (3) cross-lingual reference translation to extend English-only annotations to all target languages. The resulting corpus contains over 1M training instances spanning all six tasks and four language pairs, summarized in Table 1.
\subsection{Translation Augmentation}
\label{subsec:translation_augmentation}
Most datasets provide annotations only in English. To cover all target languages (de, it, zh), we translate English references using \texttt{translategemma-12b-it}\footnote{\huggingfacesmall{} \href{https://huggingface.co/google/translategemma-12b-it}{google/translategemma-12b-it}} \citep{gemmateam2025gemma3technicalreport}. 

We select this model based on strong reference-free translation quality (COMETKiwi \cite{rei2022cometkiwi}), outperforming alternatives such as SeamlessM4T-Large \cite{barrault2023seamlessm4t} and LLaMA-3.1-8B-Instruct \cite{kassianik2025llama}. This procedure is applied consistently across tasks for language coverage augmentation.


\subsection{ASR}
\label{section:asr}
Our original ASR datasets consist of short audio-transcript pairs, typically under 15 seconds. 
However, the long-form track requires audio up to 15 minutes, necessitating enhancements to generate training data aligned with this evaluation setting.

\paragraph{YTSeg.} 
We repurpose YTSeg \cite{retkowski-waibel-2024-text}, a dataset originally curated for chaptering, as a source of long-form video for ASR. 
We retain videos up to 15 minutes, reducing the dataset from 16{,}404 to 10{,}729 examples (34.6\% reduction).
We further filter videos where Whisper Large achieves $\geq$50\% WER to remove noisy examples, yielding 10{,}638 examples (0.86\% additional reduction).
We then applied lightweight text normalization:
removing non-lexical metadata such as background-noise markers in parentheses or square brackets (e.g., [music], (applause)), while preserving bracketed numeric content. 
We also normalized whitespace and excessive line breaks.
This produced cleaner transcripts that better reflect spoken content for ASR training and evaluation.

\paragraph{NUTSHELL.} We use NUTSHELL \citep{zufle-etal-2025-nutshell}, a dataset of ACL talk videos, which is particularly in-domain for our setting. Since NUTSHELL lacks ASR transcripts, we generate them using \texttt{parakeet-tdt-0.6b-v2}\footnote{\huggingfacesmall{} \href{https://huggingface.co/nvidia/parakeet-tdt-0.6b-v2}{nvidia/parakeet-tdt-0.6b-v2}}, applied to the audio track of each video. To align with the shared task's 15-minute duration limit, we filter out videos exceeding this threshold.

\paragraph{EuroParl.} We extend EuroParl-ST \citep{iranzo-sanchez-etal-2020-europarl} for long-form ASR by concatenating aligned speech segments into 5--10 minute chunks. Segments are grouped by session and speaker, with additional speakers included if needed. Only transcribed audio portions are concatenated, excluding untranscribed gaps, yielding clean long-form audio-transcript pairs.

\paragraph{LibriSpeech.} 
We extend LibriSpeech \citep{panayotov2015librispeech} by combining \texttt{train-clean-360} and \texttt{train-other-500} to increase acoustic diversity. Utterances are grouped by chapter and shuffled at the chapter level to mix conditions while preserving intra-chapter order. Chapters are then greedily concatenated into segments of up to 10 minutes, yielding long-form audio–text pairs aligned with the evaluation setting.

We further apply truecasing and punctuation restoration using NVIDIA NeMo’s \cite{kuchaiev2019nemo} \textit{PunctuationCapitalizationModel} (punctuation\_en\_bert), a BERT-based model \cite{devlin2019bert}, converting lowercase, unpunctuated transcripts into well-formed text.

Collectively, these augmentations transform four short-form ASR corpora into long-form training data without requiring new recordings or manual annotation, demonstrating that existing resources can be effectively repurposed for long-form settings through concatenation and normalization alone.
\subsection{Speech Translation (ST)}
We use four datasets for the ST task:

\paragraph{EuroParl-ST.} We apply the same long-form concatenation pipeline used for ASR to EuroParl-ST, generating long-form speech translations for the language pairs en--de and en--it.

\paragraph{CoVost.}  We apply a similar long-form concatenation pipeline to CoVost \citep{wang2020covost} to generate long-form speech translation pairs for en--zh.

\paragraph{NUTSHELL \& LibriSpeech.} We leverage respective transcripts and translate them to German, Italian, and Chinese.

     
\subsection{Spoken Question Answering (SQA)}
We rely on two SQA benchmarks and their translations:

\paragraph{LibriSQA.} We use the open-form question subset of LibriSQA \citep{zhao2023librisqa}, a spoken question-answering dataset built on LibriSpeech \citep{panayotov2015librispeech}. We translate questions and answers into German, Italian, and Chinese as described earlier.

\paragraph{NUTSHELL.} 
To address the lack of academic-domain coverage in LibriSQA, we adapt NUTSHELL for SQA. Using ASR transcripts, we prompt \texttt{gemma-3-12b-it}\footnote{\huggingfacesmall{} \href{https://huggingface.co/google/gemma-3-12b-it}{google/gemma-3-12b-it}} \citep{gemmateam2025gemma3technicalreport} to generate five questions per transcript: four answerable from the content and one unanswerable.

\subsection{Multiple Choice (MC)}
To improve generalization and mitigate overfitting and catastrophic forgetting, as well as to prepare for the surprise task, we incorporate multiple-choice style data.

\paragraph{LibriSQA (LibriMC).}  We use the multiple-choice subset of LibriSQA \citep{zhao2023librisqa} and translate questions into the missing three target languages.

\paragraph{MMSU.}
We leverage the Massive Multi-task Spoken Language Understanding and Reasoning Benchmark (MMSU) \cite{wang2025mmsu}, using its multiple-choice questions to strengthen instruction-following capabilities.

\subsection{Speech Summarization (SSUM)}
We use the following two datasets for SSUM, and their translations into the other target languages.

\paragraph{NUTSHELL.}
We use NUTSHELL \citep{zufle-etal-2025-nutshell}, which pairs ACL talk videos with corresponding paper abstracts. We treat abstracts as reference summaries.

\paragraph{YTSeg.}

To increase training data for the SSUM task, we augment YTSeg with synthetic abstract-like summaries. We restrict the pool to talks categorized as \textit{Science}, \textit{Technology}, or \textit{Education}, according to the LLaMA 2-based topic categories from \citet{retkowski-waibel-2024-text}, to better match the academic domain of ACL test data. This filtering step leaves 5,700 (or 53.6\%) of the 10,638 videos obtained from the preprocessing in Section \ref{section:asr} for training. We find that abstracts in NUTSHELL average approximately 145 words in length. We use this as an explicit length target when prompting \texttt{Qwen3.5-27B}\footnote{\huggingfacesmall{} \href{https://huggingface.co/Qwen/Qwen3.5-27B}{Qwen/Qwen3.5-27B}} with the reference transcript, and three randomly sampled NUTSHELL abstracts as in-context examples to generate an abstract-like summary of comparable length and style for each selected talk.


\subsection{Audio Chaptering (ACHAP)}
Finally, for audio chaptering, we include two datasets, both of which are translated into all target languages:
\paragraph{YTSeg.}
For ACHAP, we utilize YTSeg and follow the same preprocessing as described in Section~\ref{section:asr}. We frame chaptering as a structured transcription task involving joint transcription, segmentation, and title generation, effectively extending the long-form ASR task. We use Markdown formatting for the structured output.

\paragraph{NUTSHELL.}
Secondly, we augment NUTSHELL with chapter annotations generated by \texttt{Qwen3-Omni}\footnote{\huggingfacesmall{} \href{https://huggingface.co/Qwen/Qwen3-Omni-30B-A3B-Instruct}{Qwen/Qwen3-Omni-30B-A3B-Instruct}}, using the inference settings and ICL prompt from \citet{retkowski2026transcripts}, which demonstrated reasonable zero-shot performance on shorter ($\leq$20 min) single-speaker audio. We retain only samples of 3--15 minutes with WER $<30\%$ against Parakeet-Transcript and 4--11 generated chapters.


\subsection{General Instruction Following}
To further enhance generalization and preserve instruction-following capabilities across domains, we incorporate general instruction-following data.
\paragraph{TowerBlocks.}
We use TowerBlocks \cite{tower_llm_2024} to construct an augmented multimodal corpus for instruction-following fine-tuning. TowerBlocks encompasses diverse subtasks including translation, named-entity recognition, post-editing, and paraphrase generation. We focus on the UltraChat subset, which provides a context-instruction format. We first filter samples matching this structure using Llama-3.1-8B \footnote{\huggingfacesmall{} \href{https://huggingface.co/meta-llama/Llama-3.1-8B}{meta-llama/Llama-3.1-8B}}, tagging corresponding components. We then generate speech samples from the context using Kokoro-82M \footnote{\huggingfacesmall{} \href{https://huggingface.co/hexgrad/Kokoro-82M}{hexgrad/Kokoro-82M}}, a state-of-the-art TTS model. This enables the model to refer to the speech during instruction-following fine-tuning.
\begin{table}[t]
\centering
\small
\begin{tabular}{l r r r}
\toprule
\textbf{Dataset} & \textbf{\# Samples} & \textbf{Initial (\%)} & \textbf{p (\%, T=2.0)} \\
\midrule
ASR        & 19{,}248  & 1.84 & 6.12 \\
SQA        & 474{,}888 & 45.31 & 30.41 \\
MC         & 380{,}056 & 36.26 & 27.21 \\
SSUM       & 35{,}748  & 3.41 & 8.34 \\
ST         & 29{,}343  & 2.80 & 7.56 \\
AChap      & 37{,}862  & 3.61 & 8.59 \\
Instruct   & 71{,}013  & 6.78 & 11.76 \\
\midrule
\textbf{Total}        & 1{,}048{,}158 & 100.00 & 100.00 \\
\bottomrule
\end{tabular}
\caption{Training data distribution with original proportions (init) and temperature-smoothed sampling probabilities ($T=2.0$, where $p_i \propto n_i^{1/T}$).}
\label{tab:data_distribution_smoothed}
\end{table}

\section{Model Submissions}
We submit two systems: an end-to-end model as our primary submission and a  cascaded model as a contrastive system to analyze the trade-offs between the 
two approaches.
\subsection{End-To-End Model}
We select \texttt{Qwen2.5-Omni}\footnote{\huggingfacesmall{} 
\href{https://huggingface.co/Qwen/Qwen2.5-Omni-7B}{Qwen/Qwen2.5-Omni-7B}} \citep{Qwen2.5-Omni} as our primary end-to-end model and use LLamaFactory 
\cite{zheng2024llamafactory} as the training framework. We chose this model for its strong multimodal instruction-following capabilities. Due to time and 
hardware constraints, we were unable to evaluate the more recent  Qwen3-Omni \cite{xu2025qwen3omnitechnicalreport}.

\subsection{Cascaded Model}
\label{subsec:cascaded_model}
We additionally explore a cascaded pipeline that decomposes the task into ASR followed by text-based instruction following. For ASR, we use \texttt{parakeet-tdt-0.6b-v2}\footnote{\huggingfacesmall{} \href{https://huggingface.co/nvidia/parakeet-tdt-0.6b-v2}{nvidia/parakeet-tdt-0.6b-v2}}, which generates transcripts from audio. These are passed to \texttt{Qwen2.5-7B-Instruct}\footnote{\huggingfacesmall{} \href{https://huggingface.co/Qwen/Qwen2.5-7B-Instruct}{Qwen/Qwen2.5-7B-Instruct}} together with task-specific prompts to produce final outputs.

To match the shared task format, we replace the \texttt{<audio>} field with a \texttt{Transcript:} prefix followed by the ASR transcript, enabling fully text-based processing. Unless stated otherwise, we use the same training data and hyperparameters as in the end-to-end setup. For MMSU, transcripts are generated with \texttt{parakeet-tdt-0.6b-v2}. Inference uses greedy decoding (beam size 1) with a maximum length of 4096 tokens.


\section{Experimental Setup}
We now detail our training recipe,  covering prompt design,  hyperparameters, data interleaving strategies, and evaluation criteria.

\subsection{Training Strategy}
To ensure our model strictly adheres to given instructions, we employ a fixed and restrictive system prompt throughout training and inference. For better instruction generalization, we create several different instruction prompts per task and per language, and each training and development instance is randomly assigned one of these prompts. The system prompt and instruction templates are provided in \cref{tab:prompt_summary} in the Appendix.
We keep the system prompt fixed because it defines a consistent global behavior across all tasks. In contrast, task-specific instructions are varied to better approximate realistic user interactions, where the same task can be expressed through many different prompt formulations. We hypothesize that prompt variation reduces template memorization and improves instruction-following robustness under prompt distribution shifts.

We also experiment with Chain-of-Thought (CoT) style predictions by introducing special task tokens with noise initialization while keeping all other settings unchanged. The task tokens include \texttt{<|asr|>}, \texttt{<|st|>}, \texttt{<|mc|>}, \texttt{<|sqa|>}, \texttt{<|achap|>}, \texttt{<|ssum|>}, and \texttt{<|instruct|>}, along with language tokens \texttt{<|en|>}, \texttt{<|de|>}, \texttt{<|it|>}, and \texttt{<|zh|>}. For SQA, we additionally include answerability tokens \texttt{<|answerable|>} and \texttt{<|unanswerable|>}. The token order follows: task token, language token, answerability token (if applicable), and then the target output.

\subsection{Data Preparation}
Our training data exhibits significant class imbalance across tasks (see ~\cref{tab:data_distribution_smoothed}). To ensure adequate task visibility during training, we maintain datasets separately and apply interleaving strategies. We explore two approaches: (1)a manually specified fixed-probability sampling strategy designed to partially compensate for task imbalance and redistribute sampling probability from less important tasks to more important ones, and (2) temperature-scaled sampling, where probabilities are derived from dataset sizes.
We do not aim to exhaustively evaluate all possible fixed-probability sampling strategies, but rather to compare a reasonable heuristic sampling distribution with a principled temperature-scaled alternative.

For fixed-probability sampling, we assign the following probabilities: ASR (10\%), SQA (31\%), MC (9\%), SSUM (14\%), ST (13\%), ACHAP (14\%), and general instruction following (9\%).
The probabilities were selected heuristically based on dataset size and perceived task importance. In particular, we reduced the MC sampling probability because MC was treated as an auxiliary task rather than a primary evaluation target.

For temperature-scaled sampling, let $n_i$ denote the number of training instances in dataset $i$ and $N=\sum_i n_i$ the total number of instances. We obtain interleaving probabilities as:
\begin{equation}
p_i = \frac{n_i^{1/T}}{\sum_j n_j^{1/T}},
\end{equation}
with $T > 0$ interpolating between size-proportional sampling ($T=1$) and uniform sampling ($T \to \infty$). We adopt $T=2$, following prior work showing improvements on the ST task \cite{li2025exploring}, which corresponds to sampling proportional to the square root of dataset size ($p_i \propto \sqrt{n_i}$).

\subsection{Hyperparameters}
We experimented with various training configurations and found that an effective batch size of 4 with a learning rate of 1.0e-4 yielded the best results, outperforming configurations with larger batch sizes and learning rates ranging from 1.0e-5 to 2.0e-4. To prevent out-of-memory errors, we set the token cutoff length to 28,000 tokens. Since 15 minutes of audio requires approximately 22,500 tokens, this provides sufficient headroom for model predictions. We set the warmup ratio to 0.1 with a total of 60,000 update steps, and apply LoRA \cite{pham2021efficient,hu2022lora} with rank 32.

\subsection{Evaluation}
For model selection, we evaluate on the MCIF (Multimodal Crosslingual Instruction-Following) benchmark \cite{papi2026mcif}, YTSeg, and LibriMC. We additionally include an unused YTSeg validation subset with ACHAP translations for German, Italian, and Chinese. During training, we extract a small subset from each development split to monitor evaluation loss and inform checkpoint selection. We apply the following evaluation protocols per task:
\begin{itemize}
    \item \textbf{ASR, ST, SQA, SSUM}: We follow the automated evaluation protocols of \citet{papi2026mcif}.
    \item \textbf{ACHAP}: We follow \citet{retkowski2026transcripts}, employing Collar-F1 (with $\pm$3 second tolerance) to assess segmentation quality and the Global Concatenation protocol with BERTScore to measure title quality, as implemented in the \texttt{chunkseg} package\footnote{\url{https://github.com/retkowski/chunkseg}}.
    \item \textbf{MC}: We compute accuracy as $\frac{1}{N} \sum_{i=1}^N \mathbf{1}[\hat{y}_i = y_i]$, where predictions are normalized by stripping whitespace and must exactly match one of the choices \{A, B, C, D\}.
    \item \textbf{Surprise Task}: As the nature of the surprise task is unknown at training time, we train the model to identify the most similar known task and apply the corresponding behavior.
\end{itemize}

\subsection{Results}
\label{subsec:results}
\begin{table*}[t]
\centering
\small
\setlength{\tabcolsep}{3pt}
\renewcommand{\arraystretch}{1.05}
\begin{tabular}{lcc|cc|cc|cc}
\toprule
\textbf{Models} 
& \multicolumn{2}{c|}{\textbf{SQA} ($\uparrow$)}
& \multicolumn{2}{c|}{\textbf{SSUM} ($\uparrow$)}
& \multicolumn{2}{c|}{\textbf{ASR} ($\downarrow$)}
& \multicolumn{2}{c}{\textbf{ST} ($\uparrow$)} \\
\cmidrule(lr){2-3} \cmidrule(lr){4-5} \cmidrule(lr){6-7} \cmidrule(lr){8-9}
& Fix. & Mix. & Fix. & Mix. & Fix. & Mix. & Fix. & Mix. \\
\midrule
\multicolumn{9}{l}{\textbf{Baselines}} \\
\cdashlinelr{1-9}
(1) Qwen 2.5 Omni 
& \phantom{-}30.78 & \phantom{-}32.94 
& 14.21 & 17.87 
& 53.40 & 35.35 
& 68.65 & 70.79 \\
(2) Cascaded 
& \phantom{-}27.48 & \phantom{-}27.66 
& 13.31 & 13.40 
& \textbf{5.88} & \textbf{6.85} 
& \underline{80.81} & \textit{80.46} \\
\midrule
\multicolumn{9}{l}{\textbf{Fine-tuned}} \\
\cdashlinelr{1-9}
(3)  Fixed-Prob. Sampling
& \phantom{-}36.16 & \phantom{-}36.42 
& 28.77 & 28.76 
& 30.59 & 33.61 
& 75.58 & 74.90 \\
(4) Temp. Sampling ($T=2$) 
& \phantom{-}37.75 & \phantom{-}37.87 
& 28.87 & 28.99 
& 27.58 & 38.43 
& 76.10 & 75.96 \\
(5) N2 
& \textit{\phantom{-}39.98} & \textit{\phantom{-}39.98}
& \textbf{29.05} & 28.75 
& 25.98 & 30.46 
& 74.98 & 74.94 \\
(6) N2+Avg 
& \textbf{\phantom{-}40.68} & \textbf{\phantom{-}40.86} 
& 26.06 & \textit{29.33} 
& 29.36 & 34.48 
& 73.35 & 75.42 \\
\rowcolor{gray!20}
(7) N2+IT+Avg (\textbf{Primary})
& \phantom{-}\underline{40.42} & \phantom{-}\underline{40.31} 
& \textit{26.10} & \underline{29.41} 
& 37.65 & 36.88 
& 73.94 & 76.01 \\
(8) N2+IT+SHAS 
& -11.09 & -12.14 
& \phantom{0}2.93 & \phantom{0}5.90 
& \textit{11.72} & \textit{11.25} 
& \textit{80.62} & \underline{80.84} \\
(9) Chain-of-Thought (CoT) 
& \phantom{-}34.88 & \phantom{-}35.50 
& 26.56 & \textbf{29.74} 
& 79.24 & 80.04 
& 42.09 & 43.69 \\
\rowcolor{gray!20}
(10) Cascaded+FT (\textbf{Contrastive})
& \phantom{-}33.36 & \phantom{-}32.09 
& \underline{28.55} & 27.33 
& \phantom{0}\underline{5.90} & \phantom{0}\underline{9.76} 
& \textbf{83.72} & \textbf{83.75} \\
\bottomrule
\end{tabular}
\\[4pt]
\begin{minipage}{0.8\textwidth}
\footnotesize
\textbf{Legend:}
(5) continued fine-tuning on in-domain (NUTSHELL) data;
(6) checkpoint averaging of (5);
(7) additional Italian in-domain fine-tuning with averaging;
(10) cascaded system with fine-tuning.
\end{minipage}
\caption{MCIF long track comparison of fixed vs.\ mixed prompt evaluation
(macro-averaged). Best results are in \textbf{bold}, second-best are
\underline{underlined} and third-best in \textit{italic}. \colorbox{gray!20}{Shaded rows} indicate submitted models.}
\label{tab:fixed_vs_mixed}
\end{table*}
We report results on the MCIF long-form track as well as on the ACHAP and MC tasks, which are evaluated separately due to their distinct evaluation protocols.

\subsubsection{MCIF Results}
\cref{tab:fixed_vs_mixed} presents results on the MCIF long-form track with fixed and mixed prompts and beam size~1. We compare two baselines: Qwen2.5-Omni (end-to-end) and Cascaded (Section~\ref{subsec:cascaded_model}).

\paragraph{Baselines.} The cascaded baseline demonstrates the advantage of using an ASR model specifically optimized for long-form audio, achieving 5.88\% WER and 80.81\% COMET score on translation. In contrast, the end-to-end baseline outperforms the cascaded approach on SQA and SSUM, but its inferior ASR and ST performance (53.40\% WER, 68.65\% COMET) reveals that it was not specifically designed for long-form audio.

\paragraph{Fine-tuning.} Fine-tuning substantially improves both systems. We observe faster convergence and better final results with temperature-scaled data sampling (row 4) compared to our manually specified fixed-probability sampling strategy (row 3), and therefore use it for all subsequent experiments.

Temperature-scaled sampling with $T=2$ consistently performs well across tasks (Table~2, rows 3 vs.~4). As $T=2$ corresponds to sampling proportional to $\sqrt{n_i}$, this extends observations of \cite{li2025exploring} from text-only to multimodal speech instruction following.
%
This suggests that square-root proportional sampling is a strong default for multimodal speech instruction-following training.

\paragraph{In-Domain Adaptation and Checkpoint Averaging.} 
Continuing training on NUTSHELL, an in-domain dataset of scientific talks similar to the evaluation data, yields clear improvements (row 5), supporting our hypothesis that in-domain adaptation benefits long-form performance.
Checkpoint averaging of the best two checkpoints (row 6) provides further gains, whereas averaging additional checkpoints degraded performance. Since performance on Italian remained suboptimal (see Table~\ref{tab:results_fixed_prompt}), we additionally fine-tuned row (4) with Italian in-domain data and averaged the best checkpoint with those from row (6), yielding row (7). This model shows modest improvements on SSUM and ST tasks.

\paragraph{Segmented Evaluation.} To diagnose whether poor ASR results stem from task confusion rather than fundamental ASR limitations, we applied segmented evaluation (SHAS) to row (7). The degradation on SQA and SSUM is expected, but acceptable ASR performance suggests that long-form nature, rather than the task itself, drives the poor unsegmented results.

\paragraph{Chain-of-Thought Conditioning.}
We evaluate CoT-style task-token conditioning (row 9) \citep{koneru2025kit}, but observe degraded performance, especially for ASR (79.24\% WER). 

Task prediction collapses: ASR inputs are almost always misclassified as SSUM, despite only a slight data imbalance (\textasciitilde2\%). Explicit ASR predictions are rare (3 fixed, 2 mixed) and occur only when the ground truth is SQA, indicating a failure to learn task discrimination.

We hypothesize that the model defaults to SSUM because it is both slightly more frequent and structurally closer to ASR (i.e., grounded in transcript-level content), making it an easier fallback. This suggests that prefix-based task routing remains weakly grounded and prone to shortcut behavior, requiring stronger balancing or supervision to be effective.

\paragraph{Cascaded System.} Fine-tuning the cascaded system substantially improves performance on SQA (27.48 → 33.36\%) and SSUM (13.31 → 28.55\%) while preserving strong ASR performance (5.88 → 5.90\% WER). The cascaded model achieves the best overall ST result (83.72\% COMET) among all systems. These results highlight that the cascaded approach benefits from high-quality intermediate transcripts and excels at transcription-sensitive tasks. However, performance on semantic tasks such as SQA and SSUM lags the end-to-end model, likely due to error propagation from ASR transcripts and loss of audio-specific information. Additionally, training data aligned with the end-to-end setup may not be optimal for text-based instruction following, potentially limiting the cascaded model's ability to handle semantic reasoning tasks.

\paragraph{Model Selection.} Model selection was based on robustness across all tasks. Row (7) consistently ranks as the second or third best performer across metrics, closely following the top model. For the surprise task, we hypothesized that the Italian fine-tuning checkpoint, when averaged with others, could mitigate catastrophic forgetting and improve generalization \cite{ugan2025weight}. Thus, we selected row (7) as our primary submission and row (10) as a contrastive system. Although the cascaded+FT model underperforms on SQA and SSUM, its superior ASR and ST performance and qualitative analysis suggest it may handle challenging surprise task instances.

\subsubsection{ACHAP and MC Results}
Results on ACHAP and MC tasks are presented in Table~\ref{tab:mc_achap_results} in the Appendix. We evaluated MC only on English, as we identified it as a potential surprise task and reserved other languages for evaluation robustness. For ACHAP, we used 50 unused samples from the YTSeg development split, translating them into German, Italian, and Chinese for multilingual evaluation.

Both fine-tuned models substantially outperform the baseline on MC, achieving over 81\% accuracy. For ACHAP, multilingual training yields large gains over the baseline in English, improving F1 from 1.20 to over 34. 
Across all evaluated languages, both models achieve comparable performance, indicating that the multilingual augmentation strategy successfully transfers the task to Chinese, German, and Italian. 
While N2+Avg obtains the highest F1 scores on English, German, and Italian, N2+IT+Avg achieves the best MC accuracy, the highest Chinese F1 score, and the strongest BERTScores on German and Italian.

\subsection{Re-ranking}\label{sec:reranking}
To improve generation quality beyond single-pass decoding, we generate $N=17$ candidate outputs per segment: one greedy decode, one greedy decode with SHAS-based segmentation~\citep{huber2023end, tsiamas22_interspeech}, and 15 sampled candidates. We then apply re-ranking to select the best hypothesis. 

A key challenge is that the task identity is unknown at inference time: the re-ranker therefore receives audio, the original prompt instructions, and candidates. This realistic constraint prevents task-specific strategies and requires a single method to generalize across several tasks.

%
%

 \paragraph{Reranking Strategies.}
We evaluate six re-ranking strategies: \textit{Likelihood} scores each candidate independently via the model's conditional probability; \textit{Comparison} presents all candidates simultaneously and asks the model to select the best; \textit{Pairwise} runs a sequential tournament where the greedy candidate serves as champion and faces each remaining candidate in turn; \textit{Pairwise Round-Robin} compares every pair of candidates once and selects the candidate with the most wins; \textit{Pairwise Bracket} organises comparisons as a single-elimination bracket, mitigating positional bias from sequential tournaments; and \textit{MBR} selects the candidate with the highest average chrF similarity to all others, requiring no model inference. 
Additional details, including inference costs, can be found in~\cref{tab:reranking_methods}.

\begin{table}[ht]
\small
\centering
\resizebox{\columnwidth}{!}{%
\begin{tabular}{lrrrrrr}
\toprule
\textbf{Method} & \textbf{ASR} ($\downarrow$) & \textbf{SQA} ($\uparrow$) & \textbf{SSUM} ($\uparrow$) & \textbf{ST} ($\uparrow$) & \textbf{Impr.} \\
\midrule
Greedy & 40.77 & 37.24 & 29.38 & 75.28 & --- \\
Oracle & -32.10 & +14.42 & +3.85 & +6.11 & +14.12 \\
\midrule
Likelihood & \textbf{-24.93} & -11.06 & -8.60 & +0.93 & +1.55 \\
Comparison & +4.16 & -4.25 & -2.03 & -1.35 & -2.95 \\
Pairw. & -5.62 & -3.78 & -1.71 & -1.02 & -0.22 \\
Pairw. RR & -1.39 & -3.36 & -1.84 & -1.51 & -1.33 \\
Pairw. Brack. & -1.83 & -3.25 & -1.52 & -0.63 & -0.89 \\
MBR & +3.09 & \textbf{+0.22} & \textbf{-0.79} & +0.27 & -0.85 \\
Lik.\,+\,MBR & -19.28 & -3.33 & -2.19 & \textbf{+1.09} & \textbf{+3.71} \\
\cdashlinelr{1-6}
Casc. Pairw. & +3.39 & -3.32 & -2.29 & -1.83 & -2.71 \\
Casc. Pairw. RR & +14.77 & -3.66 & -1.45 & -0.65 & -5.13 \\
Casc. Pairw. Brack. & +12.43 & -2.97 & -1.76 & -1.24 & -4.60 \\
\bottomrule
\end{tabular}%
}
\caption{Re-ranking results. Values show $\Delta$ vs.\ greedy. ASR uses WER (lower $\downarrow$ is better); SQA/SSUM/ST use BERTScore/COMET (higher $\uparrow$ is better). Casc.\ = Whisper + Gemma two-stage pipeline; Pairw.\ = Pairwise; RR = Round-Robin; Lik.\,+\,MBR = Likelihood and MBR combined with a tiebreaking pairwise comparison. Greedy row shows absolute scores. Impr.\ = mean signed improvement (ASR sign-flipped).}
\label{tab:reranking_results}
\end{table}

\paragraph{Reranking Models.}
We primarily use \texttt{Qwen2.5-Omni}\footnote{\huggingfacesmall{} \href{https://huggingface.co/Qwen/Qwen2.5-Omni-7B}{Qwen/Qwen2.5-Omni-7B}}
 \citep{Qwen2.5-Omni} as the re-ranking model, since it can directly condition on the audio input. We additionally experiment with a two-stage pipeline combining \texttt{whisper-large-v3}\footnote{\huggingfacesmall{} \href{https://huggingface.co/openai/whisper-large-v3}{openai/whisper-large-v3}}
 \cite{radford2022robustspeechrecognitionlargescale} for audio transcription and \texttt{gemma-3-12b-it}\footnote{\huggingfacesmall{} \href{https://huggingface.co/google/gemma-3-12b-it}{google/gemma-3-12b-it}}
 \citep{gemmateam2025gemma3technicalreport} for text-based candidate comparison, applying all three pairwise strategies.                                                  

\subsubsection{Reranking Results}
Results averaged across languages are shown in \cref{tab:reranking_results}. The oracle upper bound reveals substantial potential gains for ASR ($-32.1$ WER points) and SQA ($+14.4$), indicating high diversity among the 17 candidates. For SSUM and ST, the oracle improvements are considerably smaller ($+3.9$ and $+2.0$ respectively), suggesting less variance and thus limited headroom for re-ranking.

In practice, no single method reliably captures these gains across all tasks. \textit{Likelihood} achieves the strongest ASR improvement ($-24.9$ WER points), but at the cost of large degradations on SQA and SSUM. \textit{MBR} is conservative: it preserves SQA quality but provides no ASR benefit.
Combining both signals via \textit{Lik.+MBR}, which resolves disagreements between Likelihood and MBR through a single pairwise comparison using the same pairwise re-ranking model, retains most of the ASR gain ($-19.3$) while substantially reducing the SQA degradation ($-2.9$), yielding the best overall score.
Per-language results (\cref{tab:reranking_results_en,tab:reranking_results_de,tab:reranking_results_it,tab:reranking_results_zh} Appendix) show consistent gains only for English and Chinese, while German and Italian show no reliable improvement over greedy.   For the final submission, we therefore apply \textit{Lik.\,+\,MBR} re-ranking only for English and Chinese.  

\subsubsection{Reranking Analysis}
We analyze two aspects of re-ranker behavior that explain the inconsistent results: positional bias and spurious SHAS-candidate selection.

\paragraph{Positional Bias.} Pairwise methods are susceptible to spurious preferences for whichever candidate appears first (A-bias) or second (B-bias) in a comparison. We quantify this as 
\begin{equation}
    P(\text{pick A} \mid \text{B better}) - 
P(\text{pick B} \mid \text{A better})
\end{equation}
where positive values indicate A-bias and negative values B-bias. 
Results are shown in \cref{tab:position_bias} in \cref{app:reranking}. Sequential \textit{Pairwise} exhibits strong A-bias ($+35.3$~pp on average, up to $+56.8$~pp for German), which is expected since the greedy candidate always starts as champion and thus always occupies the first position. \textit{Round-Robin} nearly eliminates this bias ($+3.2$~pp), since every pair is compared once in each order. \textit{Bracket} introduces a slight B-bias ($-5.5$~pp) with high variance across languages. These systematic biases likely explain the inconsistent performance of sequential and bracket methods in \cref{tab:reranking_results}.

 \paragraph{SHAS candidate selection.} A second source of error is the spurious selection of the SHAS-segmented candidate for tasks where accurate segmentation provides no benefit. As shown in \cref{tab:shas_selection} in \cref{app:reranking}, \textit{Likelihood} selects SHAS frequently for SQA ($22.7$\%) and SSUM ($34.5$\%), directly explaining its large degradations. \textit{MBR} avoids this entirely, the SHAS output tends to be less fluent and therefore scores poorly under chrF-based similarity , but consequently fails to select it for ASR and ST, where it would be beneficial. \textit{Lik.,+,MBR} strikes the best balance: spurious SHAS selection drops sharply for SQA $(7.4$\%) and SSUM ($5.5$\%), while ASR ($19.0$\%) and ST ($25.4$\%) retain it at useful rates. This explains why \textit{Lik.,+,MBR} achieves the best overall score in \cref{tab:reranking_results}.

These results suggest a general principle for multi-task re-ranking without task identity: likelihood alone is too aggressive, MBR alone is too conservative, and their combination provides the best tradeoff by using MBR as a regularizer that suppresses spurious likelihood-driven candidate selection. We recommend Lik.+MBR as a default re-ranking strategy for multi-task speech instruction-following systems where task identity is unavailable at inference time.

\subsection{Submission setting}
\begin{table}[t]
\centering
\scriptsize
\setlength{\tabcolsep}{4pt}
\renewcommand{\arraystretch}{1.05}
\begin{tabular}{lccc|ccc}
\toprule
\textbf{Method} 
& \multicolumn{3}{c|}{\textbf{EN}} 
& \multicolumn{3}{c}{\textbf{ZH}} \\
\cmidrule(lr){2-4} \cmidrule(lr){5-7} 
& SQA$\uparrow$ & SSUM$\uparrow$ & ASR$\downarrow$
& SQA$\uparrow$ & SSUM$\uparrow$ & ST$\uparrow$ \\
\midrule

(7)
& \textbf{44.91} & \textbf{23.44} & 37.65
& 39.31 & 39.57 & 79.06 \\

(7)+Reranker 
& 44.30 & 23.10 & \textbf{21.39}
& \textbf{40.52} & \textbf{40.11} & \textbf{80.61} \\

\bottomrule
\end{tabular}
\caption{MCIF-long fixed-prompt evaluation comparing N2+IT+Avg and its reranked variant. Best values per column are in \textbf{bold}.}
\label{tab:reranker_comparison}
\end{table}
We selected model (7) as our submission setup and additionally apply the re-ranking strategy from Section~\ref{sec:reranking}.
The re-ranking reduced performance on Italian and German; we therefore report only the English and Chinese results on the MCIF long track in~\cref{tab:reranker_comparison}.
While English shows slight degradation on SQA and SSUM, the ASR improvement is substantial (37.65 $\rightarrow$ 21.39 WER).
Chinese shows consistent improvements across tasks.
These results motivate submitting the re-ranking strategy for the English and Chinese tracks.

For the Short Track, we submitted model (7) as the primary system due to its strong WER on pre-segmented audio, with the transcript-based system as the contrastive submission.

\subsection{Official IWSLT Evaluation Results}
Table~\ref{tab:official_results_summary} summarizes the official IWSLT evaluation results reported by the shared-task organizers \cite{adelani-etal-2026-iwslt}, averaged across languages.
Full language-specific results are provided in Appendix Table~\ref{tab:official_iwslt_results}.
The results highlight complementary strengths of the two submissions.
In the Long Track, the contrastive system achieves the strongest ST, quality estimation (QE), the surprise task, and ASR performance, while the primary system performs better on SQA, SSUM, and ACHAP.
In contrast, the primary submission achieves the strongest Short Track results on ST, SQA, and ASR, whereas the contrastive submission substantially outperforms it on QE.
Overall, neither submission consistently outperforms the other across all tasks.

Notably, the primary submissions achieve 0.0 QE accuracy in both tracks, while the contrastive submission reaches 0.722, highlighting a substantial difference between the audio-based and transcript-based pipelines on the unseen QE task.

\begin{table}[t]
\centering
\scriptsize
\setlength{\tabcolsep}{4pt}
\renewcommand{\arraystretch}{1.05}
\begin{tabular}{lcccccc}
\hline
\textbf{Submission}
& \textbf{ST}
& \textbf{SQA}
& \textbf{QE}
& \textbf{ASR}
& \textbf{SSUM}
& \textbf{ACHAP} \\

& COMET
& BERT
& Acc.
& WER
& BERT
& F1 \\
\hline

S Primary
& \textbf{0.844}
& \textbf{0.484}
& 0.000
& \textbf{0.074}
& --
& -- \\

S Contrastive
& 0.838
& 0.448
& \textbf{0.722}
& 0.170
& --
& -- \\
\hline
L Primary
& 0.751
& \textbf{0.427}
& 0.000
& 0.269
& \textbf{0.275}
& \textbf{0.474} \\

L Contrastive
& \textbf{0.843}
& 0.344
& \textbf{0.722}
& \textbf{0.064}
& 0.268
& 0.421 \\
\hline
\end{tabular}
\caption{Official IWSLT evaluation results averaged across languages. Metrics correspond to COMET (ST), QA-BERTScore (SQA), QE accuracy (QE), WER (ASR), BERTScore (SSUM), and CollarF1 (ACHAP). S for Short Track and L for Long Track.}
\label{tab:official_results_summary}
\end{table}

\section{Conclusion}
This work presents five contributions to long-form speech instruction following: (1) a general data augmentation framework for constructing long-form training data from short-form sources; (2) an empirical comparison of fixed-probability and temperature-scaled sampling strategies; (3) a negative result on CoT task-token conditioning, revealing a collapse in task discrimination due to weak grounding and task similarity rather than simple data imbalance; (4) a systematic analysis of re-ranking under the realistic constraint of unknown task identity; and (5) a combined Lik.+MBR reranking strategy that mitigates the trade-off between transcription and semantic tasks. 

We release our data augmentation code and model checkpoints to support future research on long-form multimodal instruction following \footnote{\huggingfacesmall{}\href{https://huggingface.co/datasets/YapayNet/iwslt2026-if-augmented}{YapayNet/iwslt2026-if-augmented}}.

\section*{Acknowledgments}
This work was supported by the project ``How is AI Changing Science? Research in the Era of Learning Algorithms'' (HiAICS), funded by the Volkswagen Foundation, and partially by the European Union’s Horizon research and innovation programme under grant agreement No. 101135798, project Meetween (My Personal AI Mediator for Virtual MEETtings BetWEEN People) and European Union’s Horizon Europe programme grant agreement No. 101213369 (DVPS).
The authors gratefully acknowledge computing time provided on
HoreKa at the National High-Performance Computing Center at KIT
(NHR@KIT), supported by the Federal Ministry of Education and
Research, the Ministry of Science, Research and the Arts of
Baden-Württemberg, and the DFG.

\bibliography{custom}

@inproceedings{adelani-etal-2026-iwslt,
    title     = {Speech Translation and Metrics in 2026: Findings of the IWSLT Campaign},
    author    = {
               Adelani, David Ifeoluwa
                and Anastasopoulos, Antonios
                and Agostinelli, Victor
                and Bentivogli, Luisa
                and Bojar, Ond{\v{r}}ej
                and Brati{\`e}res, Sebastien
                and Carpuat, Marine
                and Cattoni, Roldano
                and Cettolo, Mauro
                and Chen, Lizhong
                and Federico, Marcello
                and Gaido, Marco
                and Gupta, Mahendra
                and Han, HyoJung
                and Hatami, Ali
                and Javorsk{\'y}, David
                and Jeon, Yejin
                and Kasztelnik, Marek
                and Liu, Danni
                and Luu, Nam
                and Ma, Min
                and Mach{\'a}{\v{c}}ek, Dominik
                and Maltais, Marie
                and Matusov, Evgeny
                and Maurya, Chandresh Kumar
                and McCrae, John P.
                and Moslem, Yasmin
                and Nakamura, Satoshi
                and Negri, Matteo
                and Niehues, Jan
                and Ojha, Atul Kr.
                and Ouyang, Siqi
                and Papi, Sara
                and Pol{\'a}k, Peter
                and Retkowski, Fabian
                and Savoldi, Beatrice
                and Sikasote, Claytone
                and Sperber, Matthias
                and St{\"u}ker, Sebastian
                and Sudoh, Katsuhito
                and Turchi, Marco
                and Waibel, Alex
                and Wilken, Patrick
                and Zouhar, Vil{\'e}m
                and Z{\"u}fle, Maike
                },
    booktitle = {Proceedings of the 23rd International Conference on Spoken Language Translation (IWSLT 2026)},
    year      = {2026},
    address = "San Diego, California, US",
    publisher = "Association for Computational Linguistics",
}

@inproceedings{zufle-niehues-2025-contrastive,
    title = "Contrastive Learning for Task-Independent {S}peech{LLM}-Pretraining",
    author = {Z{\"u}fle, Maike  and
      Niehues, Jan},
    editor = "Che, Wanxiang  and
      Nabende, Joyce  and
      Shutova, Ekaterina  and
      Pilehvar, Mohammad Taher",
    booktitle = "Findings of the Association for Computational Linguistics: ACL 2025",
    month = jul,
    year = "2025",
    address = "Vienna, Austria",
    publisher = "Association for Computational Linguistics",
    url = "https://aclanthology.org/2025.findings-acl.445/",
    doi = "10.18653/v1/2025.findings-acl.445",
    pages = "8469--8490",
    ISBN = "979-8-89176-256-5"
}

@inproceedings{sinhamahapatra2025slides,
  title={Do Slides Help? Multi-modal Context for Automatic Transcription of Conference Talks},
  author={Sinhamahapatra, Supriti and Niehues, Jan},
  booktitle={Proceedings of the 2025 Conference on Empirical Methods in Natural Language Processing},
  pages={16111--16121},
  year={2025}
}

@inproceedings{huber2023end,
  title={End-to-end evaluation for low-latency simultaneous speech translation},
  author={Huber, Christian and Dinh, Tu Anh and Mullov, Carlos and Pham, Ngoc-Quan and Nguyen, Thai-Binh and Retkowski, Fabian and Constantin, Stefan and Ugan, Enes and Liu, Danni and Li, Zhaolin and others},
  booktitle={Proceedings of the 2023 Conference on Empirical Methods in Natural Language Processing: System Demonstrations},
  pages={12--20},
  year={2023}
}

@article{wang2025mmsu,
      title={MMSU: A Massive Multi-task Spoken Language Understanding and Reasoning Benchmark}, 
      author={Dingdong Wang and Jincenzi Wu and Junan Li and Dongchao Yang and Xueyuan Chen and Tianhua Zhang and Helen Meng},
      journal={arXiv preprint arXiv:2506.04779},
      year={2025},
}

@inproceedings{devlin2019bert,
  title={Bert: Pre-training of deep bidirectional transformers for language understanding},
  author={Devlin, Jacob and Chang, Ming-Wei and Lee, Kenton and Toutanova, Kristina},
  booktitle={Proceedings of the 2019 conference of the North American chapter of the association for computational linguistics: human language technologies, volume 1 (long and short papers)},
  pages={4171--4186},
  year={2019}
}

@article{kuchaiev2019nemo,
  title={Nemo: a toolkit for building ai applications using neural modules},
  author={Kuchaiev, Oleksii and Li, Jason and Nguyen, Huyen and Hrinchuk, Oleksii and Leary, Ryan and Ginsburg, Boris and Kriman, Samuel and Beliaev, Stanislav and Lavrukhin, Vitaly and Cook, Jack and others},
  journal={arXiv preprint arXiv:1909.09577},
  year={2019}
}

@inproceedings{koneru2026boom,
  title={BOOM: Beyond Only One Modality KIT’s Multimodal Multilingual Lecture Companion},
  author={Koneru, Sai and Retkowski, Fabian and Huber, Christian and Hilgert, Lukas and Akti, Seymanur and Ugan, Enes Yavuz and Waibel, Alex and Niehues, Jan},
  booktitle={Proceedings of the 19th Conference of the European Chapter of the Association for Computational Linguistics (Volume 3: System Demonstrations)},
  pages={175--187},
  year={2026}
}

@article{abdin2024phi,
  title={Phi-4 technical report},
  author={Abdin, Marah and Aneja, Jyoti and Behl, Harkirat and Bubeck, S{\'e}bastien and Eldan, Ronen and Gunasekar, Suriya and Harrison, Michael and Hewett, Russell J and Javaheripi, Mojan and Kauffmann, Piero and others},
  journal={arXiv preprint arXiv:2412.08905},
  year={2024}
}

@inproceedings{tangsalmonn,
  author       = {Changli Tang and
                  Wenyi Yu and
                  Guangzhi Sun and
                  Xianzhao Chen and
                  Tian Tan and
                  Wei Li and
                  Lu Lu and
                  Zejun Ma and
                  Chao Zhang},
  title        = {{SALMONN:} Towards Generic Hearing Abilities for Large Language Models},
  booktitle    = {The Twelfth International Conference on Learning Representations,
                  {ICLR} 2024, Vienna, Austria, May 7-11, 2024},
  publisher    = {OpenReview.net},
  year         = {2024},
  url          = {https://openreview.net/forum?id=14rn7HpKVk},
  bibsource    = {dblp computer science bibliography, https://dblp.org}
}

@inproceedings{koneru2025kit,
  title={KIT’s Offline Speech Translation and Instruction Following Submission for IWSLT 2025},
  author={Koneru, Sai and Z{\"u}fle, Maike and Nguyen, Thai-Binh and Akti, Seymanur and Niehues, Jan and Waibel, Alex},
  booktitle={Proceedings of the 22nd International Conference on Spoken Language Translation (IWSLT 2025)},
  pages={232--244},
  year={2025}
}

@article{li2025exploring,
  title={Exploring the impact of temperature on large language models: Hot or cold?},
  author={Li, Lujun and Sleem, Lama and Nichil, Geoffrey and State, Radu and others},
  journal={Procedia Computer Science},
  volume={264},
  pages={242--251},
  year={2025},
  publisher={Elsevier}
}

@article{ugan2025weight,
  title={Weight factorization and centralization for continual learning in speech recognition},
  author={Ugan, Enes Yavuz and Pham, Ngoc-Quan and Waibel, Alexander},
  journal={arXiv preprint arXiv:2506.16574},
  year={2025}
}

@misc{xu2025qwen25omnitechnicalreport,
      title={Qwen2.5-Omni Technical Report}, 
      author={Jin Xu and Zhifang Guo and Jinzheng He and Hangrui Hu and Ting He and Shuai Bai and Keqin Chen and Jialin Wang and Yang Fan and Kai Dang and Bin Zhang and Xiong Wang and Yunfei Chu and Junyang Lin},
      year={2025},
      eprint={2503.20215},
      archivePrefix={arXiv},
      primaryClass={cs.CL},
      url={https://arxiv.org/abs/2503.20215}, 
}

@misc{xu2025qwen3omnitechnicalreport,
      title={Qwen3-Omni Technical Report}, 
      author={Jin Xu and Zhifang Guo and Hangrui Hu and Yunfei Chu and Xiong Wang and Jinzheng He and Yuxuan Wang and Xian Shi and Ting He and Xinfa Zhu and Yuanjun Lv and Yongqi Wang and Dake Guo and He Wang and Linhan Ma and Pei Zhang and Xinyu Zhang and Hongkun Hao and Zishan Guo and Baosong Yang and Bin Zhang and Ziyang Ma and Xipin Wei and Shuai Bai and Keqin Chen and Xuejing Liu and Peng Wang and Mingkun Yang and Dayiheng Liu and Xingzhang Ren and Bo Zheng and Rui Men and Fan Zhou and Bowen Yu and Jianxin Yang and Le Yu and Jingren Zhou and Junyang Lin},
      year={2025},
      eprint={2509.17765},
      archivePrefix={arXiv},
      primaryClass={cs.CL},
      url={https://arxiv.org/abs/2509.17765}, 
}

@INPROCEEDINGS{iranzo-sanchez-etal-2020-europarl,
  author={Iranzo-Sánchez, Javier and Silvestre-Cerdà, Joan Albert and Jorge, Javier and Roselló, Nahuel and Giménez, Adrià and Sanchis, Albert and Civera, Jorge and Juan, Alfons},
  booktitle={ICASSP 2020 - 2020 IEEE International Conference on Acoustics, Speech and Signal Processing (ICASSP)}, 
  title={Europarl-ST: A Multilingual Corpus for Speech Translation of Parliamentary Debates}, 
  year={2020},
  volume={},
  number={},
  pages={8229-8233},
  doi={10.1109/ICASSP40776.2020.9054626}}

@inproceedings{papi2026mcif,
title={{MCIF}: Multimodal Crosslingual Instruction-Following Benchmark from Scientific Talks},
author={Sara Papi and Maike Z{\"u}fle and Marco Gaido and Beatrice Savoldi and Danni Liu and Ioannis Douros and Luisa Bentivogli and Jan Niehues},
booktitle={The Fourteenth International Conference on Learning Representations},
year={2026},
url={https://openreview.net/forum?id=PtPYZYfa0h}
}

@inproceedings{rei2022cometkiwi,
  title={CometKiwi: IST-unbabel 2022 submission for the quality estimation shared task},
  author={Rei, Ricardo and Treviso, Marcos and Guerreiro, Nuno M and Zerva, Chrysoula and Farinha, Ana C and Maroti, Christine and De Souza, Jos{\'e} GC and Glushkova, Taisiya and Alves, Duarte and Coheur, Luisa and others},
  booktitle={Proceedings of the Seventh Conference on Machine Translation (WMT)},
  pages={634--645},
  year={2022}
}

@misc{gemmateam2025gemma3technicalreport,
      title={Gemma 3 Technical Report}, 
      author={Gemma Team and Aishwarya Kamath and Johan Ferret and Shreya Pathak and Nino Vieillard and Ramona Merhej and Sarah Perrin and Tatiana Matejovicova and Alexandre Ramé and Morgane Rivière and Louis Rouillard and Thomas Mesnard and Geoffrey Cideron and Jean-bastien Grill and Sabela Ramos and Edouard Yvinec and Michelle Casbon and Etienne Pot and Ivo Penchev and Gaël Liu and Francesco Visin and Kathleen Kenealy and Lucas Beyer and Xiaohai Zhai and Anton Tsitsulin and Robert Busa-Fekete and Alex Feng and Noveen Sachdeva and Benjamin Coleman and Yi Gao and Basil Mustafa and Iain Barr and Emilio Parisotto and David Tian and Matan Eyal and Colin Cherry and Jan-Thorsten Peter and Danila Sinopalnikov and Surya Bhupatiraju and Rishabh Agarwal and Mehran Kazemi and Dan Malkin and Ravin Kumar and David Vilar and Idan Brusilovsky and Jiaming Luo and Andreas Steiner and Abe Friesen and Abhanshu Sharma and Abheesht Sharma and Adi Mayrav Gilady and Adrian Goedeckemeyer and Alaa Saade and Alex Feng and Alexander Kolesnikov and Alexei Bendebury and Alvin Abdagic and Amit Vadi and András György and André Susano Pinto and Anil Das and Ankur Bapna and Antoine Miech and Antoine Yang and Antonia Paterson and Ashish Shenoy and Ayan Chakrabarti and Bilal Piot and Bo Wu and Bobak Shahriari and Bryce Petrini and Charlie Chen and Charline Le Lan and Christopher A. Choquette-Choo and CJ Carey and Cormac Brick and Daniel Deutsch and Danielle Eisenbud and Dee Cattle and Derek Cheng and Dimitris Paparas and Divyashree Shivakumar Sreepathihalli and Doug Reid and Dustin Tran and Dustin Zelle and Eric Noland and Erwin Huizenga and Eugene Kharitonov and Frederick Liu and Gagik Amirkhanyan and Glenn Cameron and Hadi Hashemi and Hanna Klimczak-Plucińska and Harman Singh and Harsh Mehta and Harshal Tushar Lehri and Hussein Hazimeh and Ian Ballantyne and Idan Szpektor and Ivan Nardini and Jean Pouget-Abadie and Jetha Chan and Joe Stanton and John Wieting and Jonathan Lai and Jordi Orbay and Joseph Fernandez and Josh Newlan and Ju-yeong Ji and Jyotinder Singh and Kat Black and Kathy Yu and Kevin Hui and Kiran Vodrahalli and Klaus Greff and Linhai Qiu and Marcella Valentine and Marina Coelho and Marvin Ritter and Matt Hoffman and Matthew Watson and Mayank Chaturvedi and Michael Moynihan and Min Ma and Nabila Babar and Natasha Noy and Nathan Byrd and Nick Roy and Nikola Momchev and Nilay Chauhan and Noveen Sachdeva and Oskar Bunyan and Pankil Botarda and Paul Caron and Paul Kishan Rubenstein and Phil Culliton and Philipp Schmid and Pier Giuseppe Sessa and Pingmei Xu and Piotr Stanczyk and Pouya Tafti and Rakesh Shivanna and Renjie Wu and Renke Pan and Reza Rokni and Rob Willoughby and Rohith Vallu and Ryan Mullins and Sammy Jerome and Sara Smoot and Sertan Girgin and Shariq Iqbal and Shashir Reddy and Shruti Sheth and Siim Põder and Sijal Bhatnagar and Sindhu Raghuram Panyam and Sivan Eiger and Susan Zhang and Tianqi Liu and Trevor Yacovone and Tyler Liechty and Uday Kalra and Utku Evci and Vedant Misra and Vincent Roseberry and Vlad Feinberg and Vlad Kolesnikov and Woohyun Han and Woosuk Kwon and Xi Chen and Yinlam Chow and Yuvein Zhu and Zichuan Wei and Zoltan Egyed and Victor Cotruta and Minh Giang and Phoebe Kirk and Anand Rao and Kat Black and Nabila Babar and Jessica Lo and Erica Moreira and Luiz Gustavo Martins and Omar Sanseviero and Lucas Gonzalez and Zach Gleicher and Tris Warkentin and Vahab Mirrokni and Evan Senter and Eli Collins and Joelle Barral and Zoubin Ghahramani and Raia Hadsell and Yossi Matias and D. Sculley and Slav Petrov and Noah Fiedel and Noam Shazeer and Oriol Vinyals and Jeff Dean and Demis Hassabis and Koray Kavukcuoglu and Clement Farabet and Elena Buchatskaya and Jean-Baptiste Alayrac and Rohan Anil and Dmitry and Lepikhin and Sebastian Borgeaud and Olivier Bachem and Armand Joulin and Alek Andreev and Cassidy Hardin and Robert Dadashi and Léonard Hussenot},
      year={2025},
      eprint={2503.19786},
      archivePrefix={arXiv},
      primaryClass={cs.CL},
      url={https://arxiv.org/abs/2503.19786}, 
}

@misc{radford2022robustspeechrecognitionlargescale,
      title={Robust Speech Recognition via Large-Scale Weak Supervision}, 
      author={Alec Radford and Jong Wook Kim and Tao Xu and Greg Brockman and Christine McLeavey and Ilya Sutskever},
      year={2022},
      eprint={2212.04356},
      archivePrefix={arXiv},
      primaryClass={eess.AS},
      url={https://arxiv.org/abs/2212.04356}, 
}

@article{finkelstein2026translategemma,
  title={TranslateGemma technical report},
  author={Finkelstein, Mara and Caswell, Isaac and Domhan, Tobias and Peter, Jan-Thorsten and Juraska, Juraj and Riley, Parker and Deutsch, Daniel and Kovacs, Geza and Dilanni, Cole and Cherry, Colin and others},
  journal={arXiv preprint arXiv:2601.09012},
  year={2026}
}

@inproceedings{tsiamas22_interspeech,
  author={Ioannis Tsiamas and Gerard I. Gállego and José A. R. Fonollosa and Marta R. Costa-jussà},
  title={{SHAS: Approaching optimal Segmentation for End-to-End Speech Translation}},
  year=2022,
  booktitle={Proc. Interspeech 2022},
  pages={106--110},
  doi={10.21437/Interspeech.2022-59}
}

@inproceedings{kumar-byrne-2004-minimum,
    title = "Minimum {B}ayes-Risk Decoding for Statistical Machine Translation",
    author = "Kumar, Shankar  and
      Byrne, William",
    booktitle = "Proceedings of the Human Language Technology Conference of the North {A}merican Chapter of the Association for Computational Linguistics: {HLT}-{NAACL} 2004",
    month = may # " 2 - " # may # " 7",
    year = "2004",
    address = "Boston, Massachusetts, USA",
    publisher = "Association for Computational Linguistics",
    url = "https://aclanthology.org/N04-1022/",
    pages = "169--176"
}

@article{Qwen2.5-Omni,
  title={Qwen2.5-Omni Technical Report},
  author={Jin Xu and Zhifang Guo and Jinzheng He and Hangrui Hu and Ting He and Shuai Bai and Keqin Chen and Jialin Wang and Yang Fan and Kai Dang and Bin Zhang and Xiong Wang and Yunfei Chu and Junyang Lin},
  journal={arXiv preprint arXiv:2503.20215},
  year={2025}
}

@article{kassianik2025llama,
  title={Llama-3.1-foundationai-securityllm-base-8b technical report},
  author={Kassianik, Paul and Saglam, Baturay and Chen, Alexander and Nelson, Blaine and Vellore, Anu and Aufiero, Massimo and Burch, Fraser and Kedia, Dhruv and Zohary, Avi and Weerawardhena, Sajana and others},
  journal={arXiv preprint arXiv:2504.21039},
  year={2025}
}

@article{barrault2023seamlessm4t,
  title={Seamlessm4t: Massively multilingual \& multimodal machine translation},
  author={Barrault, Lo{\"\i}c and Chung, Yu-An and Meglioli, Mariano Cora and Dale, David and Dong, Ning and Duquenne, Paul-Ambroise and Elsahar, Hady and Gong, Hongyu and Heffernan, Kevin and Hoffman, John and others},
  journal={arXiv preprint arXiv:2308.11596},
  year={2023}
}

@article{pham2021efficient,
  title={Efficient weight factorization for multilingual speech recognition},
  author={Pham, Ngoc-Quan and Nguyen, Tuan-Nam and St{\"u}ker, Sebastian and Waibel, Alexander},
  journal={arXiv preprint arXiv:2105.03010},
  year={2021}
}

@article{hu2022lora,
  title={Lora: Low-rank adaptation of large language models.},
  author={Hu, Edward J and Shen, Yelong and Wallis, Phillip and Allen-Zhu, Zeyuan and Li, Yuanzhi and Wang, Shean and Wang, Liang and Chen, Weizhu and others},
  journal={Iclr},
  volume={1},
  number={2},
  pages={3},
  year={2022}
}

@inproceedings{zheng2024llamafactory,
  title={LlamaFactory: Unified Efficient Fine-Tuning of 100+ Language Models},
  author={Yaowei Zheng and Richong Zhang and Junhao Zhang and Yanhan Ye and Zheyan Luo and Zhangchi Feng and Yongqiang Ma},
  booktitle={Proceedings of the 62nd Annual Meeting of the Association for Computational Linguistics (Volume 3: System Demonstrations)},
  address={Bangkok, Thailand},
  publisher={Association for Computational Linguistics},
  year={2024},
  url={http://arxiv.org/abs/2403.13372}
}

@article{zhao2023librisqa,
         title={LibriSQA: Advancing Free-form and Open-ended Spoken Question Answering with a Novel Dataset and Framework},
         author={Zhao, Zihan and Jiang, Yiyang and Liu, Heyang and Wang, Yanfeng and Wang, Yu},
         journal={arXiv preprint arXiv:2308.10390},
         year={2023}
}

@misc{wang2020covost,
    title={CoVoST 2: A Massively Multilingual Speech-to-Text Translation Corpus},
    author={Changhan Wang and Anne Wu and Juan Pino},
    year={2020},
    eprint={2007.10310},
    archivePrefix={arXiv},
    primaryClass={cs.CL}
}

@inproceedings{panayotov2015librispeech,
  title={Librispeech: an asr corpus based on public domain audio books},
  author={Panayotov, Vassil and Chen, Guoguo and Povey, Daniel and Khudanpur, Sanjeev},
  booktitle={2015 IEEE international conference on acoustics, speech and signal processing (ICASSP)},
  pages={5206--5210},
  year={2015},
  organization={IEEE}
}

@inproceedings{zufle-etal-2025-nutshell,
    title = "{NUTSHELL}: A Dataset for Abstract Generation from Scientific Talks",
    author = {Z{\"u}fle, Maike  and
      Papi, Sara  and
      Savoldi, Beatrice  and
      Gaido, Marco  and
      Bentivogli, Luisa  and
      Niehues, Jan},
    editor = "Salesky, Elizabeth  and
      Federico, Marcello  and
      Anastasopoulos, Antonis",
    booktitle = "Proceedings of the 22nd International Conference on Spoken Language Translation (IWSLT 2025)",
    month = jul,
    year = "2025",
    address = "Vienna, Austria (in-person and online)",
    publisher = "Association for Computational Linguistics",
    url = "https://aclanthology.org/2025.iwslt-1.2/",
    doi = "10.18653/v1/2025.iwslt-1.2",
    pages = "19--32",
    ISBN = "979-8-89176-272-5"
}

@inproceedings{retkowski-waibel-2024-text,
    title = "From Text Segmentation to Smart Chaptering: A Novel Benchmark for Structuring Video Transcriptions",
    author = "Retkowski, Fabian  and
      Waibel, Alexander",
    editor = "Graham, Yvette  and
      Purver, Matthew",
    booktitle = "Proceedings of the 18th Conference of the European Chapter of the Association for Computational Linguistics (Volume 1: Long Papers)",
    month = mar,
    year = "2024",
    address = "St. Julian{'}s, Malta",
    publisher = "Association for Computational Linguistics",
    url = "https://aclanthology.org/2024.eacl-long.25/",
    doi = "10.18653/v1/2024.eacl-long.25",
    pages = "406--419"
}

@misc{retkowski2026transcripts,
      title={Beyond Transcripts: A Renewed Perspective on Audio Chaptering}, 
      author={Fabian Retkowski and Maike Züfle and Thai Binh Nguyen and Jan Niehues and Alexander Waibel},
      year={2026},
      eprint={2602.08979},
      archivePrefix={arXiv},
      primaryClass={cs.SD},
      url={https://arxiv.org/abs/2602.08979}, 
}

@misc{tower_llm_2024,
      title={Tower: An Open Multilingual Large Language Model for Translation-Related Tasks}, 
      author={Duarte M. Alves and José Pombal and Nuno M. Guerreiro and Pedro H. Martins and João Alves and Amin Farajian and Ben Peters and Ricardo Rei and Patrick Fernandes and Sweta Agrawal and Pierre Colombo and José G. C. de Souza and André F. T. Martins},
      year={2024},
      eprint={2402.17733},
      archivePrefix={arXiv},
      primaryClass={cs.CL}
}

@inproceedings{retkowski-etal-2025-summarizing,
    title = "Summarizing Speech: A Comprehensive Survey",
    author = {Retkowski, Fabian  and
      Z{\"u}fle, Maike  and
      Sudmann, Andreas  and
      Pfau, Dinah  and
      Watanabe, Shinji  and
      Niehues, Jan  and
      Waibel, Alexander},
    editor = "Christodoulopoulos, Christos  and
      Chakraborty, Tanmoy  and
      Rose, Carolyn  and
      Peng, Violet",
    booktitle = "Proceedings of the 2025 Conference on Empirical Methods in Natural Language Processing",
    month = nov,
    year = "2025",
    address = "Suzhou, China",
    publisher = "Association for Computational Linguistics",
    url = "https://aclanthology.org/2025.emnlp-main.1388/",
    doi = "10.18653/v1/2025.emnlp-main.1388",
    pages = "27275--27306",
    ISBN = "979-8-89176-332-6"
}
\clearpage
\appendix
\onecolumn
\crefalias{section}{appendix}
\crefalias{subsection}{appendix}
\crefalias{subsubsection}{appendix}

\setcounter{table}{0}
\renewcommand{\thetable}{A\arabic{table}}

\setcounter{figure}{0}
\renewcommand{\thefigure}{A\arabic{figure}}

\section{Appendix}
\section{Data}
\label{app:data}
\paragraph{Prompts \& Instructions}
We employ an LLM to generate paraphrased variants of manually designed instruction prompts. 
For readability, we report a representative subset in Table~\ref{tab:prompt_summary}.
\begin{table*}[t]
\centering
\scriptsize
\setlength{\tabcolsep}{3pt}
\renewcommand{\arraystretch}{1.05}
\begin{tabular}{llc>{\raggedright\arraybackslash}p{10.7cm}}
\hline
\multicolumn{4}{l}{\textbf{System Prompts}} \\
\multicolumn{4}{p{15.3cm}}{\raggedright
(1) You are a helpful assistant specialized in audio and video processing tasks. You follow instructions exactly and don't provide additional explanations or follow-up questions.\newline
(2) You are a helpful assistant specialized in audio processing tasks. You follow instructions exactly and don't provide additional explanations or follow-up questions.
} \\
\hline
\textbf{Task} & \textbf{Lang.} & \textbf{\#Instr.} & \textbf{Representative instruction variants} \\
\hline

\textbf{ASR} & EN & 30 &
(1) Transcribe the entire audio recording into plain text.\\
& & & (2) Generate a faithful transcript of this audio segment. \\
& & & (3) Convert this audio into text exactly as spoken. \\
\hline

\multirow{3}{*}{\textbf{ST}} 
& DE & 10 &
(1) Konvertieren Sie dieses Audio in deutschen Text.\\
& & & (2) Anhören und ins Deutsche übersetzen.\\
& & & (3) Was wird gesagt? Geben Sie die deutsche Übersetzung an. \\

& IT & 10 &
(1) Converti questo audio in testo italiano.\\
& & &(2) Ascolta e traduci in italiano. \\
& & & (3) Cosa viene detto? Fornisci la traduzione in italiano. \\

& ZH & 10 &
(1) \zh{请将其翻译成中文。} \\ 
& & &(2) \zh{请提供这段录音的中文翻译。} \\
& & & (3) \zh{翻译此音频为中文。} \\
\hline

\multirow{4}{*}{\textbf{ACHAP}}
& EN & 10 &
(1) Organize this recording into meaningful chapters. For each one, include a Markdown heading (\# Title) enclosed by two newlines (\texttt{\textbackslash n\textbackslash n}), followed by its transcript.\\
& & & (2)Break this audio into coherent sections as chapters. For every chapter, write a Markdown heading (\# Title) with two newlines around it (\texttt{\textbackslash n\textbackslash n}), then add the transcript for that part.\\
& & &(3) Create coherent chapter segments for this audio. Each chapter should have a Markdown heading (\# Title) with two surrounding newlines (\texttt{\textbackslash n\textbackslash n}), then the transcript of that chapter.\\

& DE & 10 &
(1) Unterteile dieses Audio in zusammenhaengende Kapitel und transkribiere jedes davon. Beginne jedes Kapitel mit einer Markdown-Ueberschrift (\# Title), umgeben von zwei Zeilenumbruechen (\texttt{\textbackslash n\textbackslash n}), und fuege dann das Transkript an.\\
& & & (2)Segmentiere dieses Audio in zusammenhaengende Kapitel. Fuer jedes Kapitel gib eine Markdown-Ueberschrift (\# Title) mit zwei umgebenden Zeilenumbruechen (\texttt{\textbackslash n\textbackslash n}) aus und fuege danach das Transkript dieses Abschnitts an.\\
& & &(3) Teile diese Aufnahme in sinnvolle Kapitel auf. Fuer jedes Kapitel schreibe eine Markdown-Ueberschrift (\# Title), die von zwei Zeilenumbruechen (\texttt{\textbackslash n\textbackslash n}) umgeben ist, und danach den Kapiteltext. \\

& IT & 10 &
(1) Dividi questo audio in capitoli logici. Ogni capitolo deve iniziare con un'intestazione Markdown (\# Title) circondata da due nuove righe (\texttt{\textbackslash n\textbackslash n}), seguita dal testo trascritto.\\
& & & (2) Suddividi questa registrazione in capitoli coerenti. Per ogni capitolo, inserisci un'intestazione Markdown (\# Title) con due nuove righe attorno (\texttt{\textbackslash n\textbackslash n}), poi aggiungi la trascrizione del capitolo.\\
& & & (3) Segmenta questa registrazione seguendo i confini dei capitoli. Per ogni capitolo, usa un'intestazione Markdown (\# Title) avvolta da due nuove righe (\texttt{\textbackslash n\textbackslash n}), poi fornisci la trascrizione. \\

& ZH & 10 &
(1) \zh{请将这段音频按逻辑分成多个章节。每章以 Markdown 标题（\# Title）开头，标题前后各有两个换行 (\texttt{\textbackslash n\textbackslash n})，并在后面附上对应转写文本。}\\ 
& & &(2) \zh{ 请按章节边界对这段录音进行切分。每个章节使用 Markdown 标题（\# Title），标题前后各两个换行 (\texttt{\textbackslash n\textbackslash n})，然后提供该章节转写。} \\
& & &(3) \zh{请把这段音频章节化为连贯部分。每个部分都输出 Markdown 标题（\# Title），标题前后各两个换行 (\texttt{\textbackslash n\textbackslash n})，然后写该部分转写。} \\
\hline

\multirow{4}{*}{\textbf{MC}}
& EN & 10 &
(1) Reply with only the choice. \\
& & &(2) Reply with ``The answer is'' followed by the choice.\\
& & & (3) Reply with ``Option'' followed by the choice. \\

& DE & 10 &
(1) Antworte nur mit der Auswahl.\\
& & &(2) Antworte mit „Die Antwort ist“ gefolgt von der Auswahl.\\
& & & (3) Antworte mit „Option“ gefolgt von der Auswahl. \\

& IT & 10 &
(1) Rispondi solo con la scelta. \\
& & &(2) Rispondi con ``La risposta è'' seguito dalla scelta.\\
& & &(3) Rispondi con ``Opzione'' seguito dalla scelta. \\

& ZH & 10 &
(1) \zh{只回答选项字母。}\\ 
& & &(2) \zh{请回答“答案是”加上选项字母。}\\
& & &(3) \zh{请回答“选项”加上选项字母。} \\
\hline

\multirow{4}{*}{\textbf{SQA}}
& EN & 10 &
(1) Based on the English content, respond to this question with a brief answer:\\
& & & (2) Use the English content to provide a concise answer to the question below:\\
& & &(3) Refer to the English content to answer the question. Be concise: \\

& DE & 20 &
(1) Beantworte die Frage basierend auf dem Inhalt kurz:\\
& & &(2) Nutze den Inhalt und antworte prägnant auf die folgende Frage:\\
& & &(3) Beantworte die Frage basierend auf dem englischen Inhalt kurz: \\

& IT & 20 &
(1) In base al contenuto, rispondi brevemente alla domanda:\\
& & &(2) Usa il contenuto per fornire una risposta concisa:\\
& & &(3) In base al contenuto inglese, rispondi brevemente alla domanda: \\

& ZH & 20 &
(1) \zh{请基于内容，简要回答这个问题：}\\
& & &(2) \zh{根据内容，对下列问题给出简洁回答：}\\
& & &(3) \zh{请基于英文内容，简要回答这个问题：} \\
\hline

\multirow{4}{*}{\textbf{SSUM}}
& EN & 10 &
(1) Summarize the English audio.\\
& & &(2) Provide a concise summary of the English audio.\\
& & &(3) Summarize the English audio in at most \{x\} words. \\

& DE & 10 &
(1) Fassen Sie das englische Audio zusammen.\\
& & &(2) Erstellen Sie eine knappe Zusammenfassung des englischen Audios.\\
& & &(3) Fassen Sie das englische Audio in maximal \{x\} Wörtern zusammen. \\

& IT & 10 &
(1) Riassumi l'audio inglese.\\
& & &(2) Fornisci un riassunto conciso dell'audio in inglese.\\
& & &(3) Riassumi l'audio inglese in massimo \{x\} parole. \\

& ZH & 10 &
(1) \zh{概括英文音频内容。}\\
& & &(2) \zh{请简要总结英文音频的主要内容。}\\
& & &(3) \zh{请将英文音频内容概括为不超过\{x\}个词。} \\
\hline
\end{tabular}
\caption{Compact summary of the instruction templates used in training after deduplication. We report the number of unique templates per task/language and show three representative examples for each setting. For NUTSHELL data we used System Prompt (1) as and the others (2).In SSUM, the \{x\} was replaced with the number of words in the target summery.}
\label{tab:prompt_summary}
\end{table*}

\clearpage
\section{Reranking Strategies}
\label{app:reranking}
\paragraph{Setting.}    We generate $N=17$ candidates per segment: one greedy decode, one greedy
  decode with SHAS-based segmentation~\citep{tsiamas22_interspeech}, and 15   
  samples (temperature $0.8$).                                                
  Candidates are produced by our fine-tuned model, while re-ranking is        
  performed by the base \texttt{Qwen2.5-Omni-7B}~\citep{Qwen2.5-Omni}, which  
  can directly condition on the audio input.
  For the cascaded methods, audio is first transcribed with \texttt{whisper-large-v3}~\citep{radford2022robustspeechrecognitionlargescale} and the        
  resulting transcript is passed together with all candidates to
  \texttt{gemma-3-12b-it}~\citep{gemmateam2025gemma3technicalreport} for      
  text-based comparison.                                 
  An overview of the re-ranking strategies and the number of model calls
  required for each is given in \cref{tab:reranking_methods}.       
  
      \begin{table*}[ht]                                                              \small                                
  \centering                                                                                                                           
                                                                         
  \begin{tabular}{p{1.5cm}p{10cm}c}                                                                                
  \toprule                                                                                                 
  \textbf{Method} & \textbf{Description} & \textbf{Model calls} \\
  \midrule                                                                                                 
  Likelihood & Each candidate is scored independently by computing $\log p(\text{candidate} \mid           
  \text{audio})$ with an external model. The candidate with the highest likelihood is selected. & $N$ \\
  \addlinespace                                                                                            
  Comparison & The model receives the audio, the task, and all candidates simultaneously and is prompted to identify
  the best candidate. & $1$ \\                                                                           
  \addlinespace                           
  Pairwise & A sequential tournament: candidate~0 (greedy) acts as champion and is compared against candidates~1    
  through $N{-}1$ in order; the winner of each comparison advances. & $N{-}1$ \\                           
  \addlinespace                           
  Pairwise Round-Robin & Every pair of candidates is compared once; the candidate with the most wins is    
  selected. Unbiased with respect to candidate order. & $\frac{N(N-1)}{2}$ \\                              
  \addlinespace                           
  Pairwise Bracket & A single-elimination bracket where winners are matched against winners across $\log_2 
  N$ rounds, avoiding the positional bias of the sequential tournament. & $N{-}1$ \\                       
  \addlinespace                           
  MBR & Minimum Bayes Risk~\citep{kumar-byrne-2004-minimum}: the candidate with the highest average chrF similarity 
  to all other candidates is selected. & $0$ \\                                                            
  \bottomrule                             
  \end{tabular}                                                                                                                 
  \caption{Overview of re-ranking methods. For model-based methods, the original instruction is included in the prompt of the reranker since the task is determined at inference time.}       \label{tab:reranking_methods}             
  \end{table*}

  \paragraph{Per-language results.} We use the MCIF test set   \citep{papi2026mcif} to evaluate our re-ranking methods using the official  MCIF evaluation code.                                     
  Results per language are shown in \cref{tab:reranking_results_en} (en),     
  \cref{tab:reranking_results_de} (de), \cref{tab:reranking_results_it} (it), 
  and \cref{tab:reranking_results_zh} (zh).
  Re-ranking yields consistent improvements only for English, where Likelihood
   achieves the largest gain, driven primarily by a strong improvement on ASR.
  For Chinese, modest improvements are observed with Likelihood and MBR.
  For German and Italian, no method reliably improves over greedy across all  
  tasks.     

\begin{table}[ht]
\centering
\begin{tabular}{lrrrr}
\toprule
\textbf{Method} & \textbf{ASR} $\downarrow$ & \textbf{SQA} $\uparrow$ & \textbf{SSUM} $\uparrow$ & \textbf{Impr.} \\
\midrule
Greedy & 40.77 & 38.43 & 23.43 & --- \\
Oracle & -32.10 & +14.84 & +5.51 & +17.48 \\
\midrule
Likelihood & \textbf{-24.93} & -9.21 & -2.52 & +4.40 \\
Comparison & +4.16 & -4.63 & -2.10 & -3.63 \\
Pairw. & -5.62 & -3.19 & -1.94 & +0.16 \\
Pairw. RR & -1.39 & -3.37 & -2.07 & -1.35 \\
Pairw. Brack. & -1.83 & -2.67 & -1.32 & -0.72 \\
MBR & +3.09 & \textbf{+1.59} & \textbf{-0.41} & -0.64 \\
Lik.\,+\,MBR & -19.28 & -2.94 & -2.18 & \textbf{+4.72} \\
\addlinespace
Casc. Pairw. & +3.39 & -2.46 & -2.28 & -2.71 \\
Casc. Pairw. RR & +14.77 & -3.74 & -1.50 & -6.67 \\
Casc. Pairw. Brack. & +12.43 & -1.98 & -1.46 & -5.29 \\
\bottomrule
\end{tabular}%

\caption{Re-ranking results --- English. Values show $\Delta$ vs.\ greedy. ASR uses WER (lower $\downarrow$ is better); QA/SUM/ST use BERTScore/COMET (higher $\uparrow$ is better). Casc.\ = Whisper + Gemma two-stage pipeline; Pairw.\ = Pairwise; RR = Round-Robin; Lik.\,+\,MBR = Likelihood and MBR combined with a tiebreaking pairwise comparison. Greedy row shows absolute scores. Impr.\ = mean signed improvement (ASR sign-flipped). \textbf{Bold}: best method per column.}
\label{tab:reranking_results_en}
\end{table}


\begin{table}[ht]
\small
\centering
\begin{tabular}{lrrrr}
\toprule
\textbf{Method} & \textbf{SQA} $\uparrow$ & \textbf{SSUM} $\uparrow$ & \textbf{ST} $\uparrow$ & \textbf{Impr.} \\
\midrule
Greedy & 36.77 & 25.60 & 72.91 & --- \\
Oracle & +15.76 & +4.06 & +5.49 & +8.44 \\
\midrule
Likelihood & -14.29 & -8.01 & \textbf{+2.77} & -6.51 \\
Comparison & -2.55 & -2.55 & -1.47 & -2.19 \\
Pairw. & -2.68 & -1.09 & -1.22 & -1.66 \\
Pairw. RR & -2.26 & -1.22 & -1.05 & -1.51 \\
Pairw. Brack. & -1.79 & \textbf{-0.86} & -1.25 & -1.30 \\
MBR & \textbf{+0.70} & -1.03 & -0.46 & \textbf{-0.26} \\
\addlinespace
Casc. Pairw. & -1.98 & -2.05 & -1.46 & -1.83 \\
Casc. Pairw. RR & -2.98 & -1.42 & -0.91 & -1.77 \\
Casc. Pairw. Brack. & -1.69 & -1.74 & -1.51 & -1.65 \\
\bottomrule
\end{tabular}%

\caption{Re-ranking results --- German. Values show $\Delta$ vs.\ greedy. ASR uses WER (lower $\downarrow$ is better); QA/SUM/ST use BERTScore/COMET (higher $\uparrow$ is better). Casc.\ = Whisper + Gemma two-stage pipeline; Pairw.\ = Pairwise; RR = Round-Robin. Greedy row shows absolute scores. Impr.\ = mean signed improvement (ASR sign-flipped). \textbf{Bold}: best method per column.}
\label{tab:reranking_results_de}
\end{table}


\begin{table}[ht]
\small
\centering
\begin{tabular}{lrrrr}
\toprule
\textbf{Method} & \textbf{SQA} $\uparrow$ & \textbf{SSUM} $\uparrow$ & \textbf{ST} $\uparrow$ & \textbf{Impr.} \\
\midrule
Greedy & 35.60 & 28.38 & 74.80 & --- \\
Oracle & +13.76 & +2.72 & +6.27 & +7.58 \\
\midrule
Likelihood & -16.13 & -8.78 & \textbf{+1.19} & -7.91 \\
Comparison & -5.05 & -2.26 & -3.06 & -3.46 \\
Pairw. & -4.32 & -2.53 & -2.22 & -3.02 \\
Pairw. RR & -4.46 & -2.23 & -1.96 & -2.88 \\
Pairw. Brack. & -4.99 & -2.66 & -1.72 & -3.12 \\
MBR & \textbf{-1.49} & \textbf{-1.74} & -0.33 & \textbf{-1.19} \\
Lik.\,+\,MBR & -5.03 & -3.21 & +0.43 & -2.60 \\
\addlinespace
Casc. Pairw. & -2.80 & -3.18 & -4.70 & -3.56 \\
Casc. Pairw. RR & -3.22 & -2.39 & -1.80 & -2.47 \\
Casc. Pairw. Brack. & -2.51 & -2.52 & -3.56 & -2.86 \\
\bottomrule
\end{tabular}%

\caption{Re-ranking results --- Italian. Values show $\Delta$ vs.\ greedy. ASR uses WER (lower $\downarrow$ is better); QA/SUM/ST use BERTScore/COMET (higher $\uparrow$ is better). Casc.\ = Whisper + Gemma two-stage pipeline; Pairw.\ = Pairwise; RR = Round-Robin; Lik.\,+\,MBR = Likelihood and MBR combined with a tiebreaking pairwise comparison. Greedy row shows absolute scores. Impr.\ = mean signed improvement (ASR sign-flipped). \textbf{Bold}: best method per column.}
\label{tab:reranking_results_it}
\end{table}


\begin{table}[ht]
\small
\centering
\begin{tabular}{lrrrr}
\toprule
\textbf{Method} & \textbf{SQA} $\uparrow$ & \textbf{SSUM} $\uparrow$ & \textbf{ST} $\uparrow$ & \textbf{Impr.} \\
\midrule
Greedy & 38.17 & 40.09 & 78.12 & --- \\
Oracle & +13.32 & +3.11 & +6.56 & +7.66 \\
\midrule
Likelihood & -4.61 & -15.10 & -1.16 & -6.96 \\
Comparison & -4.78 & -1.20 & +0.48 & -1.83 \\
Pairw. & -4.94 & -1.28 & +0.38 & -1.95 \\
Pairw. RR & --- & --- & --- & --- \\
Pairw. Brack. & -3.55 & -1.24 & +1.09 & -1.23 \\
MBR & \textbf{+0.09} & \textbf{+0.01} & +1.61 & \textbf{+0.57} \\
Lik.\,+\,MBR & +0.04 & -0.43 & \textbf{+1.65} & +0.42 \\
\addlinespace
Casc. Pairw. & -6.03 & -1.65 & +0.68 & -2.33 \\
Casc. Pairw. RR & -4.68 & -0.47 & +0.77 & -1.46 \\
Casc. Pairw. Brack. & -5.69 & -1.31 & +1.35 & -1.88 \\
\bottomrule
\end{tabular}%

\caption{Re-ranking results --- Chinese. Values show $\Delta$ vs.\ greedy. ASR uses WER (lower $\downarrow$ is better); QA/SUM/ST use BERTScore/COMET (higher $\uparrow$ is better). Casc.\ = Whisper + Gemma two-stage pipeline; Pairw.\ = Pairwise; RR = Round-Robin; Lik.\,+\,MBR = Likelihood and MBR combined with a tiebreaking pairwise comparison. Greedy row shows absolute scores. Impr.\ = mean signed improvement (ASR sign-flipped). \textbf{Bold}: best method per column.}
\label{tab:reranking_results_zh}
\end{table}

\paragraph{Analysis.}
\cref{tab:position_bias} reports positional bias for each pairwise re-ranking method and language, measured as 
\begin{equation}
    P(\text{pick A} \mid \text{B better}) - 
P(\text{pick B} \mid \text{A better}).
\end{equation} Positive values indicate a spurious preference for the first candidate, negative values for the second. \cref{tab:shas_selection} reports the proportion of segments for which each re-ranking method selects the SHAS-segmented candidate, broken down by task. Both tables complement the analysis in \cref{sec:reranking}.

\begin{table*}[ht]
\small
\centering
\begin{tabular}{lrrrrr}
\toprule
\textbf{Method} & \textbf{en} & \textbf{de} & \textbf{it} & \textbf{zh} & \textbf{Avg.\ Bias} \\
\midrule
Comparison & 8.5\,/\,16 (last) & 7.4\,/\,16 (first) & 7.7\,/\,16 (first) & 7.6\,/\,16 (first) & 7.8\,/\,16 (first) \\
\midrule
Pairw. & $+$30.0 (A) & $+$56.8 (A) & $+$49.3 (A) & $+$5.1 (A) & $+$35.3 (A) \\
Pairw. RR & $+$2.3 (A) & $+$4.0 (A) & $+$3.4 (A) & --- & $+$3.2 (A) \\
Pairw. Brack. & $-$8.6 (B) & $+$11.3 (A) & $+$1.9 (A) & $-$26.8 (B) & $-$5.5 (B) \\
\addlinespace
Casc. Pairw. & $+$19.0 (A) & $+$20.8 (A) & $+$17.8 (A) & $+$10.6 (A) & $+$17.1 (A) \\
Casc. Pairw. RR & $+$1.1 (A) & $+$3.6 (A) & $+$2.5 (A) & $+$2.1 (A) & $+$2.3 (A) \\
Casc. Pairw. Brack. & $-$21.2 (B) & $-$21.5 (B) & $-$24.9 (B) & $-$29.1 (B) & $-$24.2 (B) \\
\bottomrule
\end{tabular}

\caption{Positional preference of re-ranking methods. For pairwise methods: position bias $= P(\text{pick A} \mid \text{B oracle better}) - P(\text{pick B} \mid \text{A oracle better})$, with preferred position in parentheses (A = first, B = second); values near 0 indicate no bias. For \textit{Comparison}: average selected candidate index (0 = greedy, 16 = SHAS), where 8 would be unbiased. Casc.\ = Whisper + Gemma cascade; Pairw.\ = Pairwise; RR = Round-Robin.}
\label{tab:position_bias}
\end{table*}

\begin{table}[ht]
\small
\centering
\begin{tabular}{lrrrr}
\toprule
\textbf{Method} & \textbf{ASR} & \textbf{SQA} & \textbf{SSUM} & \textbf{ST} \\
\midrule
Likelihood & 19.0 & 22.7 & 34.5 & 36.5 \\
MBR & 0.0 & 0.0 & 0.0 & 0.0 \\
Lik.\,+\,MBR & 19.0 & 7.4 & 5.5 & 25.4 \\
\bottomrule
\end{tabular}%

\caption{Rate at which the SHAS-segmented candidate is selected per task (\%). Ideally the re-ranker should prefer SHAS for ASR and ST (where accurate segmentation helps) but not for QA and SUM.}
\label{tab:shas_selection}
\end{table}


\section{Speech  Translation}
We perform English to German, Italian, and Chinese speech translation using \texttt{
translategemma-12b-it} \cite{finkelstein2026translategemma}. 
Input speech is first transcribed into English text and embedded within a structured conversational format, from which the relevant response segments are extracted for translation. To efficiently handle large-scale datasets, we employ a streaming JSON parser (ijson), enabling memory-efficient sequential processing.

Inference is conducted in batches of size 8, with the model loaded in bfloat16 precision. Inputs are formatted using the model-specific chat template and tokenized with dynamic padding before being transferred to the appropriate device. Decoding is performed using deterministic (greedy) generation with a maximum output length depending on the dataset. To improve generation stability and mitigate repetition artifacts, we apply a repetition penalty around 1.1 and enforce a 5-gram repetition constraint.

\clearpage
\section{Results}
\paragraph{Per language evaluatoins}
While our main results (Section~\ref{subsec:results}) present language-averaged scores, Tables~\ref{tab:results_fixed_prompt} and~\ref{tab:results_mixed_prompt} provide per-language breakdowns for all four languages on the MCIF long-form track with fixed and mixed prompts respectively. Table~\ref{tab:mc_achap_results} shows MC results (English only) and ACHAP results (all languages). These analyses reveal substantial performance variation across languages, with some configurations (e.g., row 7) generalizing robustly while others (e.g., row 5) excel on specific languages. This granular view helps identify how language-specific challenges—morphological complexity, data scarcity, and pre-training bias—affect different tasks and languages.

\begin{table}[h!]
\centering
\scriptsize
\setlength{\tabcolsep}{3pt}
\renewcommand{\arraystretch}{1.05}
\begin{threeparttable}
\begin{tabular}{llccc|ccc|ccc|ccc}
\hline
& & \multicolumn{3}{c|}{\textbf{EN}} 
& \multicolumn{3}{c|}{\textbf{ZH}} 
& \multicolumn{3}{c|}{\textbf{DE}} 
& \multicolumn{3}{c}{\textbf{IT}} \\
\textbf{ID} & \textbf{Method}
& SQA & SSUM & ASR
& SQA & SSUM & ST
& SQA & SSUM & ST
& SQA & SSUM & ST \\
\hline

\multicolumn{14}{l}{\textbf{Baselines}} \\
\hline

(1) & Omni 
& 24.93 & 19.34 & 53.40
& 43.03 & 9.59  & 75.88
& 28.34 & 11.55 & 61.16
& 26.82 & 16.35 & 68.92 \\

(2) & Cascaded 
& 25.07 & 19.36 & 5.88
& 26.68 & 10.31 & 84.66
& 31.64 & 11.15 & 76.55
& 26.53 & 12.41 & 81.23 \\
\hline

\multicolumn{14}{l}{\textbf{Fine-tuned}} \\
\hline

(3) & Temp 
& 37.75 & 23.03 & 30.59
& 36.83 & 39.43 & 78.24
& 35.47 & 24.86 & 73.42
& 34.58 & 27.75 & 75.08 \\

(4) & T=2 
& 41.46 & 22.10 & 27.58
& 38.77 & 39.64 & 80.20
& 35.22 & 25.58 & 73.35
& 35.53 & 28.14 & 74.75 \\

(5) & N2 
& 44.41 & 23.95 & 25.98
& 41.51 & 39.44 & 78.20
& 37.63 & 25.47 & 72.33
& 36.36 & 27.32 & 74.40 \\

(6) & N2+Avg 
& 45.00 & 23.85 & 29.36
& 39.82 & 26.23 & 72.42
& 39.82 & 26.23 & 72.42
& 38.09 & 27.93 & 75.21 \\

(7) & N2+IT+Avg 
& 44.91 & 23.44 & 37.65
& 39.64 & 26.30 & 73.10
& 39.64 & 26.30 & 73.10
& 37.49 & 28.36 & 75.62 \\

(8) & N2+IT+Avg+SHAS 
& -19.48 & 0.87 & 11.72
& 2.63 & -2.78 & 84.32
& -10.50 & 5.25 & 77.83
& -16.99 & 8.38 & 79.72 \\

(9) & CoT 
& 34.49 & 24.29 & 79.24
& 35.06 & 26.70 & 39.97
& 35.06 & 26.70 & 39.97
& 34.92 & 28.53 & 46.33 \\

(10) & Cascaded+FT 
& 33.02 & 22.02 & 5.90
& 38.50 & 39.82 & 85.00
& 31.57 & 23.74 & 81.75
& 30.36 & 28.62 & 84.42 \\
\hline
\end{tabular}

\begin{tablenotes}[flushleft]
\footnotesize
\item \textbf{Legend:} (3) temperature-based sampling; (4) temperature-based sampling with $T=2$; (5) continued fine-tuning on in-domain NUTSHELL data after (4); (6) checkpoint averaging of (5); (7) continued fine-tuning on Italian NUTSHELL data with checkpoint averaging; (8) (7) with SHAS; (9) chain-of-thought prompting; (10) cascaded system with fine-tuning.
\end{tablenotes}

\caption{MCIF-long fixed-prompt evaluation results across tasks (SQA, SSUM, ASR, ST) and languages.}
\label{tab:results_fixed_prompt}
\end{threeparttable}
\end{table}

\begin{table}[h!]
\centering
\scriptsize
\setlength{\tabcolsep}{3pt}
\renewcommand{\arraystretch}{1.05}
\begin{threeparttable}
\begin{tabular}{llccc|ccc|ccc|ccc}
\hline
& & \multicolumn{3}{c|}{\textbf{EN}} 
& \multicolumn{3}{c|}{\textbf{ZH}} 
& \multicolumn{3}{c|}{\textbf{DE}} 
& \multicolumn{3}{c}{\textbf{IT}} \\
\textbf{ID} & \textbf{Method}
& SQA & SSUM & ASR
& SQA & SSUM & ST
& SQA & SSUM & ST
& SQA & SSUM & ST \\
\hline

\multicolumn{14}{l}{\textbf{Baselines}} \\
\hline

(1) & Omni 
& 33.15 & 17.57 & 35.35
& 39.59 & 21.25 & 75.74
& 30.37 & 16.04 & 66.61
& 28.66 & 16.63 & 70.01 \\

(2) & Cascaded 
& 26.97 & 18.12 & 6.85
& 28.87 & 12.51 & 84.40
& 28.06 & 11.08 & 76.70
& 26.72 & 11.88 & 80.28 \\

\hline
\multicolumn{14}{l}{\textbf{Fine-tuned}} \\
\hline

(3) & Temp 
& 39.37 & 22.67 & 33.61
& 36.60 & 39.81 & 78.72
& 35.35 & 24.75 & 72.84
& 34.34 & 27.79 & 73.15 \\

(4) & T=2 
& 42.30 & 22.18 & 38.43
& 38.38 & 39.51 & 79.60
& 34.87 & 26.16 & 73.45
& 35.92 & 28.10 & 74.82 \\

(5) & N2 
& 44.46 & 23.35 & 30.46
& 41.34 & 39.26 & 78.35
& 38.20 & 25.12 & 72.92
& 35.93 & 27.25 & 73.56 \\

(6) & N2+Avg 
& 45.15 & 24.27 & 34.48
& 40.80 & 39.37 & 79.60
& 40.25 & 26.07 & 72.71
& 37.23 & 27.62 & 73.96 \\

(7) & N2+IT+Avg 
& 45.15 & 23.75 & 36.88
& 40.70 & 39.50 & 79.11
& 38.33 & 26.13 & 73.39
& 37.04 & 28.25 & 75.54 \\

(8) & N2+IT+SHAS 
& -17.74 & 0.74 & 11.25
& -0.95 & 7.48 & 84.25
& -13.47 & 6.89 & 77.98
& -16.39 & 8.47 & 80.30 \\

(9) & CoT 
& 37.29 & 24.06 & 80.04
& 35.30 & 39.88 & 44.82
& 34.74 & 26.56 & 41.03
& 34.66 & 28.44 & 45.22 \\

(10) & Cascaded FT 
& 32.49 & 22.06 & 9.76
& 37.62 & 38.03 & 85.19
& 29.40 & 21.68 & 81.58
& 28.83 & 27.56 & 84.47 \\
\hline
\end{tabular}

\begin{tablenotes}[flushleft]
\footnotesize
\item \textbf{Legend:} (3) temperature-based sampling; (4) temperature-based sampling with $T=2$; (5) continued fine-tuning on in-domain NUTSHELL data after (4); (6) checkpoint averaging of (5); (7) continued fine-tuning on Italian NUTSHELL data with checkpoint averaging; (8) (7) with SHAS; (9) chain-of-thought prompting; (10) cascaded system with fine-tuning.
\end{tablenotes}

\caption{MCIF-long mixed-prompt evaluation results across tasks (SQA, SSUM, ASR, ST) and languages.}
\label{tab:results_mixed_prompt}
\end{threeparttable}
\end{table}

\begin{table}[h!]
\centering
\setlength{\tabcolsep}{4pt}
\renewcommand{\arraystretch}{1.05}
\begin{tabular}{l c | cc|cc|cc|cc}
\hline
\textbf{Method} 
& \textbf{MC (EN)}$\uparrow$
& \multicolumn{2}{c|}{\textbf{EN (ACHAP)}}
& \multicolumn{2}{c|}{\textbf{ZH}}
& \multicolumn{2}{c|}{\textbf{DE}}
& \multicolumn{2}{c}{\textbf{IT}} \\
& Acc
& F1 & BERT
& F1 & BERT
& F1 & BERT
& F1 & BERT \\
\hline

Baseline 
& 0.00 
& 1.20 & 86.60
& -- & --
& -- & --
& -- & -- \\

(6) N2+Avg 
& 81.53 
& \textbf{35.86} & \textbf{87.84}
& 10.11 & \textbf{62.71}
& \textbf{45.46} & 69.58
& \textbf{40.21} & 69.37 \\

(7) N2+IT+Avg 
& \textbf{81.83}
& 34.39 & 86.75
& \textbf{10.63} & 62.15
& 44.94 & \textbf{69.95}
& 39.11 & \textbf{69.84} \\

\hline
\end{tabular}
\caption{MC accuracy (EN only) and ACHAP performance across languages. For ACHAP, we report F1 and BERTScore (GC). The baseline is only evaluated on English due to poor performance on other languages. Best values per column are in \textbf{bold}.}
\label{tab:mc_achap_results}
\end{table}

\begin{table*}[t]
\centering
\scriptsize
\setlength{\tabcolsep}{4pt}
\renewcommand{\arraystretch}{1.05}
\begin{tabular}{lllcccccc ccccc}
\hline

&
&
&
\textbf{ST}
&
\textbf{SQA}
&
\multicolumn{2}{c}{\textbf{QE}}
&
\textbf{ASR}
&
\textbf{SSUM}
&
\multicolumn{4}{c}{\textbf{ACHAP}}
\\

\cline{6-7}
\cline{10-13}

\textbf{Track}
&
\textbf{Submission}
&
\textbf{Lang}
&
COMET
&
BERT
&
Acc.
&
Fmt.
&
WER
&
BERT
&
WER
&
COMET
&
F1
&
BERT
\\
\hline

\multicolumn{13}{l}{\textbf{Short Track}}\\
\hline

Short & Primary     & EN
& -- & 0.495 & -- & -- & 0.074 & -- & -- & -- & -- & --
\\

Short & Primary     & ZH
& 0.852 & 0.456 & 0.000 & 0.000 & -- & -- & -- & -- & -- & --
\\

Short & Primary     & DE
& 0.840 & 0.466 & 0.000 & 0.000 & -- & -- & -- & -- & -- & --
\\

Short & Primary     & IT
& 0.841 & 0.519 & -- & -- & -- & -- & -- & -- & -- & --
\\

\hline

Short & Contrastive & EN
& -- & 0.450 & -- & -- & 0.170 & -- & -- & -- & -- & --
\\

Short & Contrastive & ZH
& 0.845 & 0.471 & 0.739 & 0.993 & -- & -- & -- & -- & -- & --
\\

Short & Contrastive & DE
& 0.830 & 0.423 & 0.705 & 1.000 & -- & -- & -- & -- & -- & --
\\

Short & Contrastive & IT
& 0.830 & 0.449 & -- & -- & -- & -- & -- & -- & -- & --
\\

\hline

\multicolumn{13}{l}{\textbf{Long Track}}\\
\hline

Long & Primary & EN
& -- & 0.412 & -- & -- & 0.269 & 0.212 & 0.311 & -- & 0.436 & 0.869
\\

Long & Primary & ZH
& 0.789 & 0.452 & 0.000 & 0.000 & -- & 0.383 & -- & 0.785 & 0.503 & 0.685
\\

Long & Primary & DE
& 0.733 & 0.405 & 0.000 & 0.000 & -- & 0.238 & -- & 0.740 & 0.500 & 0.690
\\

Long & Primary & IT
& 0.732 & 0.439 & -- & -- & -- & 0.267 & -- & 0.737 & 0.456 & 0.703
\\

\hline

Long & Contrastive & EN
& -- & 0.385 & -- & -- & 0.064 & 0.218 & 0.093 & -- & 0.583 & 0.877
\\

Long & Contrastive & ZH
& 0.847 & 0.360 & 0.739 & 0.993 & -- & 0.378 & -- & 0.451 & 0.103 & 0.511
\\

Long & Contrastive & DE
& 0.840 & 0.308 & 0.705 & 1.000 & -- & 0.208 & -- & 0.836 & 0.508 & 0.676
\\

Long & Contrastive & IT
& 0.841 & 0.322 & -- & -- & -- & 0.269 & -- & 0.842 & 0.489 & 0.709
\\

\hline
\end{tabular}
\caption{Official IWSLT 2026 evaluation results for our submitted systems. ST is evaluated using COMET, SQA using QA-BERTScore, QE using accuracy and format accuracy, ASR using WER, and SSUM using BERTScore. For ACHAP, we report WER, COMET, CollarF1, and GC-BERTScore.}
\label{tab:official_iwslt_results}
\end{table*}

\end{document}